\documentclass[11pt, a4paper, singlecolumn, copyright, goog]{google}

\usepackage[authoryear, sort&compress, round]{natbib}
\newcounter{noteZZctr} 

\newcounter{noteWuctr} 

\usepackage[most]{tcolorbox}
\usepackage{booktabs}
\usepackage{graphicx}
\usepackage{subcaption}
\usepackage{tikz}
\usepackage{makecell}
\usepackage{fontawesome5}
\usepackage{xcolor}
\usepackage{hyperref}
\usepackage{wrapfig}

\usepackage{listings}
\usepackage{xcolor}
\usepackage{pgfplots}
\usepgfplotslibrary{groupplots}
\definecolor{promptbg}{gray}{0.97}
\definecolor{promptgray}{gray}{0.45}

\lstdefinestyle{promptstyle}{
    basicstyle=\ttfamily\scriptsize,
    backgroundcolor=\color{promptbg},
    commentstyle=\color{promptgray},
    numbers=left,
    numberstyle=\tiny\color{promptgray},
    numbersep=6pt,
    breaklines=true,
    breakatwhitespace=false,
    showstringspaces=false,
    keepspaces=true,
    columns=fullflexible,
    tabsize=2,
    frame=single,
    rulecolor=\color{black!20},
    xleftmargin=1.5em,
    framexleftmargin=1.2em
}

\usepackage{listings}
\usepackage{xcolor}

\definecolor{codegray}{gray}{0.45}
\definecolor{codebg}{gray}{0.97}

\lstdefinestyle{appendixcode}{
    basicstyle=\ttfamily\scriptsize,
    backgroundcolor=\color{codebg},
    commentstyle=\color{codegray},
    numbers=left,
    numberstyle=\tiny\color{codegray},
    numbersep=6pt,
    breaklines=true,
    breakatwhitespace=false,
    showstringspaces=false,
    keepspaces=true,
    columns=fullflexible,
    tabsize=2,
    frame=single,
    rulecolor=\color{black!20},
    xleftmargin=1.5em,
    framexleftmargin=1.2em
}

\newtcolorbox{taskbox}[1]{
  enhanced,
  colback=black!2,
  colframe=black!25,
  boxrule=0.6pt,
  arc=1.5mm,
  left=2.2mm,
  right=2.2mm,
  top=1.8mm,
  bottom=1.8mm,
  title={\textbf{#1}},
  coltitle=black,
  colbacktitle=black!5,
  title style={font=\small}
}

\uselogo{} 

\title{Dream-RSI: Recursive Self-Improvement through Evolving Worlds}

\correspondingauthor{xidongwu@google.com, zzhangx@google.com}

\renewcommand{\today}

\author[1,2]{Tong Zheng}
\author[1]{Xidong Wu}
\author[1]{Zheng Zhang}
\author[3]{Zhankui He}
\author[1]{Chaoyi Zhang}
\author[3]{Benjamin Coleman}
\author[1]{Ruoqiao Wei}
\author[3]{Di Bai}
\author[4]{Haolin Liu}
\author[2]{Rui Liu}
\author[1]{Xue Wang}
\author[1]{Yue Zhuan}
\author[3]{Wang-Cheng Kang}
\author[1]{Renkai Xiang}
\author[2]{Heng Huang}
\author[1]{Xinwu Cheng}
\author[1]{Yunsong Guo}

\affil[1]{\thepa{}{}}
\affil[2]{University of Maryland, College Park}
\affil[3]{Google Deepmind}
\affil[4]{University of Virginia}
\begin{abstract}
Recursive self-improvement is becoming essential for autonomous AI agents, whose progress depends on discovering high-value solutions across complex domains. Effective exploration drives this process, yet managing and improving exploration strategies remains a major bottleneck. Current systems face a fundamental dilemma: fixed strategies fail to adapt as search spaces scale, while online policy optimization must navigate vast meta-search spaces under delayed, expensive feedback from long-horizon rollouts. We introduce \textsc{Dream-RSI}, a framework for scalable, recursively self-improving exploration. A lightweight orchestration layer makes exploration explicit and programmable while leaving the underlying base agent unchanged. Our key insight is that accumulated discovery history can act as a replay simulator over the realized search space. By dreaming within this simulator built from historical discovery trees, \textsc{Dream-RSI} obtains immediate, low-cost off-policy feedback to evaluate and refine exploration policies without repeated, expensive online evaluation. The improved policy is then redeployed online to drive further discovery, continuously expanding the simulator pool in a self-improving loop. Across 9 tasks in 4 domains, \textsc{Dream-RSI}
achieves competitive  quality and improves discovery efficiency in several settings.

\par\vspace{0.6em}
{\centering
\href{https://github.com/zhengkid/Dream-RSI}{%
\raisebox{-0.12em}{\Large\color{black}\faGithub}\hspace{0.35em}%
{\small\ttfamily\color{blue}github.com/zhengkid/Dream-RSI}%
}
\hspace{1.5em}\textcolor{gray}{|}\hspace{1.5em}% 分隔符
\href{https://dream-rsi.com/}{%
\raisebox{-0.12em}{\Large\color{black}\faGlobe}\hspace{0.35em}%
{\small\ttfamily\color{blue}dream-rsi.com}%
}
\par}
\end{abstract}

\begin{document}

\maketitle

\section{Introduction}
Recursive self-improvement (RSI) has emerged as an ambitious goal for autonomous AI systems~\citep{202608.0051}. A common mechanism underlying RSI is an iterative loop wherein agents propose candidate solutions, evaluate outcomes, and fold the resulting feedback into subsequent iterations. Such loops have driven substantial progress across algorithm design~\citep{novikov2025alphaevolve,romera2024mathematical}, open-ended mathematical optimization~\citep{georgiev2025mathematical,anthropic2026riemann}, systems design~\citep{jaber2026autokernel,cao2026k}, and agent self-improvement~\citep{zhang2026darwin,zhang2026hyperagents,lee2026meta,zheng2026llms}. As agents grow stronger and self-improvement targets grow harder, discovery demands long-horizon exploration over vast search spaces, often spanning thousands of proposal-evaluation cycles~\citep{ye2026evaluation,openai2026navierstokes}. At this scale, orchestrating exploration becomes critical~\citep{zheng2026parallel}: poor exploration wastes computation and time, sharply limiting the efficiency and scalability of RSI. 

Existing approaches largely rely on manually designed exploration strategies that stay fixed throughout discovery~\citep{novikov2025alphaevolve,yan2026pacevolve,du2026mlevolve,jiang2026deltaevolve,ye2026evaluation}. Such strategies cannot learn from accumulated experience and may repeatedly spend computation on unproductive search directions. Recent work therefore attempts to optimize exploration policies online~\citep{liu2026evox}, but this faces two fundamental bottlenecks. First, meta-level feedback is delayed and expensive. Unlike evaluating a single candidate, assessing an exploration policy requires observing how it shapes discovery over many proposal--evaluation cycles. Second, the meta-policy space is vast, so many alternatives must be tried before a good one is found. Together, these make meta-level improvement costly. Each policy demands a long online rollout before yielding useful signal, making the self-improvement loop hard to close at the exploration layer.

\begin{figure}
    \centering
    \vspace{-1em}\includegraphics[width=0.8\linewidth]{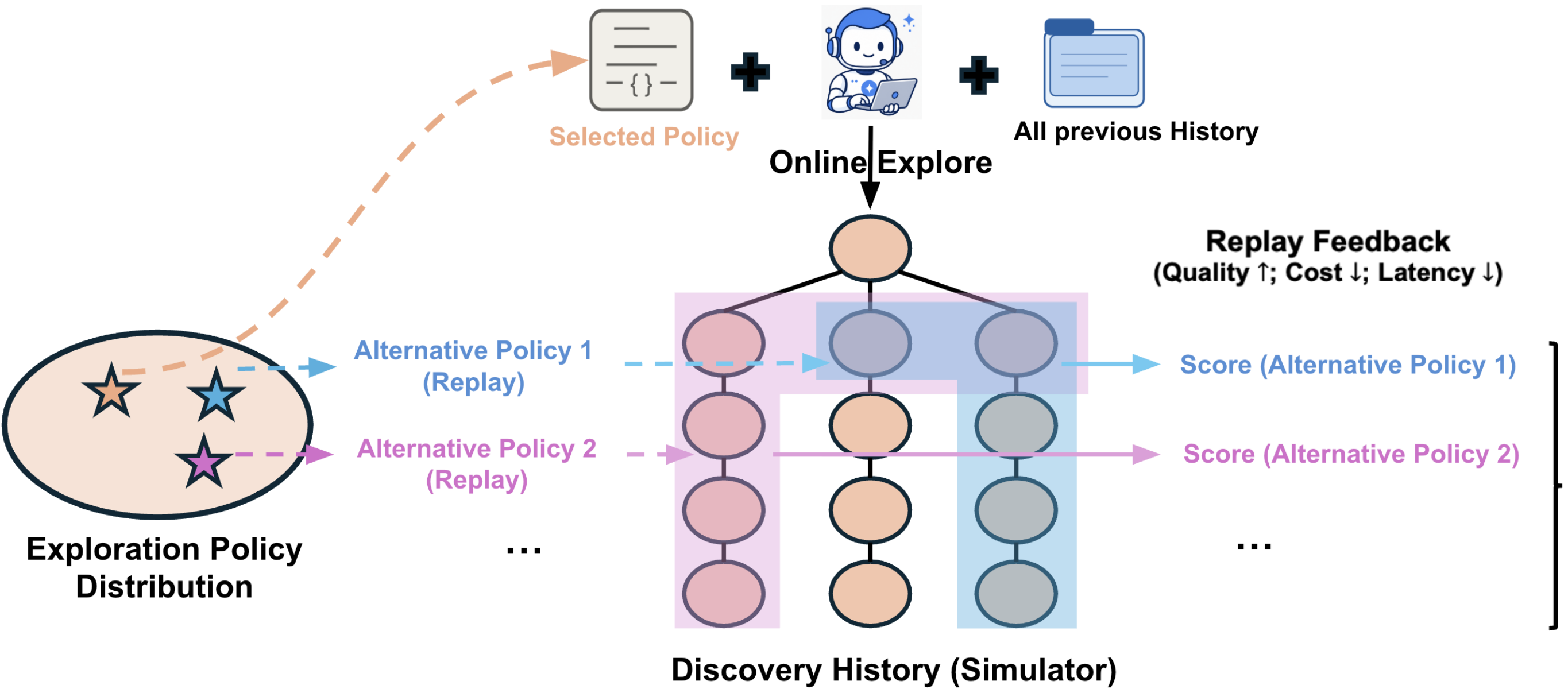}
    \caption{\textbf{Discovery history as a replay simulator.} 
    % A deployed policy first explores online to generate a structured discovery tree containing historical execution traces (each node denote an attempt with its full observation). Thousands of candidate policies can then be tested within this simulator—evaluating alternative choices of search branches, exploration orders, concurrency levels, and stopping rules. \textbf{Since all node outcomes are pre-stored, a single costly online run enables thousands of rapid, zero-execution-cost off-policy evaluations. This enables policy improvement through historical replay: the agent can “dream” over many alternative exploration strategies before redeploying the improved policy online.
    A deployed policy first explores online, recording its execution traces as a discovery tree where each node is one attempt plus its full observation. This tree then acts as a simulator: a large number of candidate policies can be tested inside it, varying which branches to search, in what order, with how much concurrency, and when to stop. \textbf{Because every node's outcome is already stored, one expensive online run can yield a large number of fast, low-cost off-policy evaluations.} The agent can therefore "dream" over many alternative exploration strategies via historical replay, and redeploy only the improved policy.}
% \vspace{-1em}

    \label{fig:motivation}
\end{figure}

To address these bottlenecks, we want to design a fast and cheap simulator of discovery to prune exploration policies before costly online deployment. \textbf{Surprisingly, completed discovery histories already provide such a simulator.} Prior work treats past history merely as static textual context~\citep{hu2025memory,ouyang2026reasoningbank} or as training data for fine-tuning~\citep{yuksekgonul2026learning,wang2025thetaevolve}. Yet a completed discovery process inherently records a structured tree of exploration decisions and their realized code-execution outcomes. By analogy to model-based reinforcement learning and World Models~\citep{ha2018world,hafner2023mastering} (\S\ref{sec:history_environment}), this discovery history can serve as a \textbf{replay simulator}\footnote{We use \emph{replay simulator} and \emph{world} interchangeably.}. Figure~\ref{fig:motivation} shows an alternative strategy can navigate this pre-recorded tree, traversing different subsets of branches in different orders, with different parallel groupings and stopping decisions. Since all execution results are locally stored, evaluating a new strategy requires only reading past records. %—no rerunning of the discovery agent or evaluator. 
This turns meta-policy improvement from expensive online trial-and-error into a fast, simulation-based ``dreaming'' procedure.

\begin{figure}
    \centering
    \vspace{-1em}\includegraphics[width=0.87\linewidth]{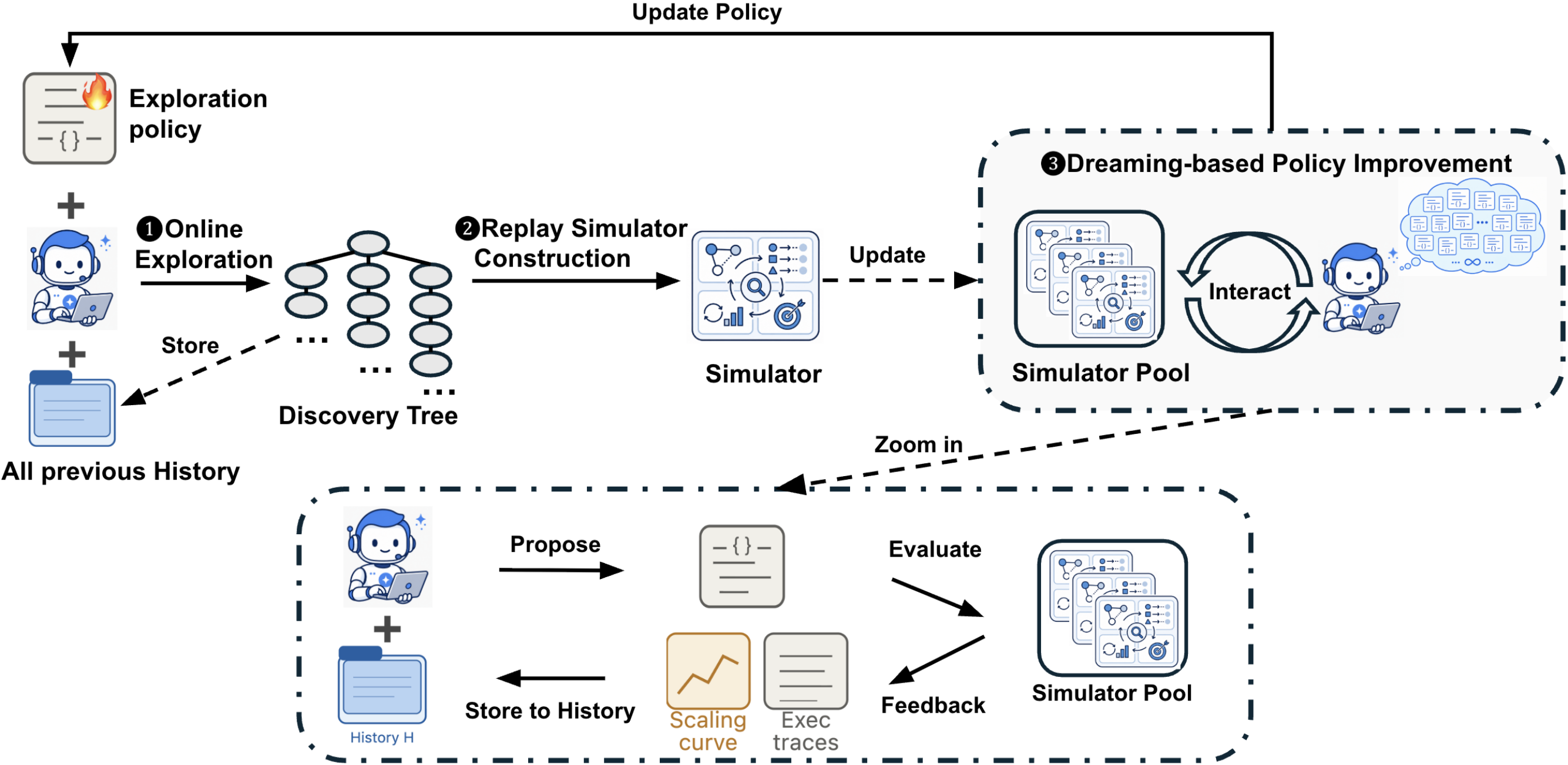}
    \caption{\textbf{Overview of \textsc{Dream-RSI}.} The system runs a recursive self-improvement loop. \textbf{\textcircled{1} Online Exploration}: the current policy drives a base agent to expand a discovery tree, logging its execution traces. \textbf{\textcircled{2} Replay Simulator Construction}: the discovery tree is distilled into a pool of reusable simulators. \textbf{\textcircled{3} Dreaming-based Policy Improvement}: the same base agent \emph{dreams}, generating many candidate policies internally and replaying them in the simulator for fast, low-cost feedback (see the zoom-in box). The refined policy is redeployed for the next round of exploration.}
    % \vspace{-1em}
    \label{fig:method}
\end{figure}

Building on this insight, we introduce \textsc{Dream-RSI}, a framework for scalable, recursively self-improving meta-exploration in agent-driven discovery. We first make exploration programmable through a lightweight orchestration layer that controls branching, parallelism, and stopping while leaving the underlying base agent unchanged. To improve this policy, \textsc{Dream-RSI} follows a self-improvement loop (Figure~\ref{fig:method}): \textbf{(1) Online Exploration}, where the current policy drives real-world discovery and logs execution traces; \textbf{(2) Replay Simulator Construction}, where recorded discovery trees become a reusable replay simulator pool; and \textbf{(3) Dreaming-based Policy Improvement}, where candidate policies are cheaply evaluated by dreaming over the simulator. The updated policy is redeployed online, generating new experience that expands the simulator pool and closing an RSI loop at the meta-exploration layer. We evaluate \textsc{Dream-RSI} on 9 discovery tasks across four domains. On \emph{algorithm engineering} (Lasso path solver), it outperforms \texttt{sklearn} and strong baselines while using up to $162\times$ fewer agent calls than \textsc{SimpleTES}. On mathematical optimization, it matches or exceeds the best results among the compared baselines on two of three tasks. On \emph{GPU kernel engineering} (KernelBench), it reaches similar speeds with $1.79\times$--$2.43\times$ fewer generations, or improves kernel performance by up to $2.09\times$ under a comparable budget. On \emph{harness optimization}, it improves the seed harness by over 8 points on Terminal-Bench 2.1.

Our contributions are: \textbf{1) History as Replay Simulator.} We recast completed discovery histories as replay simulators, making delayed exploration feedback reusable for cheap meta-exploration policy evaluation. \textbf{2) Meta-Layer RSI Loop.} We introduce \textsc{Dream-RSI}, to our knowledge the first framework to close a meta-layer RSI loop using discovery history as a replay simulator. \textbf{3) Empirical Validation.} Across nine tasks in four domains, \textsc{Dream-RSI}
achieves competitive discovery quality and improves
discovery-agent-call efficiency in several settings.

\section{Motivation: Discovery History as a Replay Simulator}
% \section{Motivation: Discovery History as a Replayable Environment}
\label{sec:history_environment}
Consider an agent navigating an unfamiliar environment. On its first attempt it wanders, hits dead ends, and backtracks — but in doing so it builds a map. That map is reusable: a new policy can plan over it, skip known dead ends, revisit earlier choices, and compare routes without physically walking them again~\citep{gupta2017cognitive}.
This is the logic of model-based reinforcement learning~\citep{SUTTON1990216, m2023model}. A model of how the environment responds to actions lets policies be trained or evaluated in simulation instead of through repeated real interaction~\citep{ha2018world}. The Dreamer family~\citep{hafner2019dream, hafner2020mastering, hafner2023mastering, hafner2025training} is the clearest example: learn a compact dynamics model from collected experience, then improve the policy by imagining trajectories inside it.

Long-horizon discovery has the same structure. An exploration policy decides which directions to pursue, which candidates to refine, what to branch on in parallel, and when to stop. Running it online leaves behind a structured history — branches explored, decisions made, costs incurred, outcomes observed. As shown in Figure~\ref{fig:motivation}, we treat this history as an empirical replay simulator: a grounded model of the part of the discovery space already seen.
Inside this simulator, a different exploration policy simply traces a different path through the recorded discovery tree, choosing other branches, another ordering, a different degree of parallelism, or an earlier stopping point. Scoring that path only requires revealing outcomes already stored along it; no rerun of the coding agent or evaluator is needed. One expensive online run therefore funds many cheap evaluations of alternative strategies.

% Requires: \usepackage{amsmath,amssymb}

\section{\textsc{Dream-RSI}: RSI through Evolving Worlds}
\label{sec:replay-based-controller-improvement}

As shown in Figure~\ref{fig:method}, \textsc{Dream-RSI} alternates between online exploration and offline ``dreaming'' for policy evaluation to improve the \emph{exploration policy} that allocates discovery computation. For online rollout, the exploration policy directs a fixed \emph{discovery agent} to generate a tree-structure discovery trace, where each node in the tree is an discovery attempt and is scored by a fixed \emph{evaluator}. The \emph{discovery tree} then becomes a \emph{replay world}: alternative policies can be re-evaluated against stored outcomes. In the offline dreaming phase, a fixed agent iteratively revises the code for the exploration policy, and every revision is evaluated in all stored replay worlds. The best version is then deployed for the next online rollout. Only the policy code changes; the models, evaluator, and execution interfaces stay fixed.

\paragraph{Discovery trees and the shared decision interface.} A discovery tree is rooted at \(r\), the initial workspace state. Every non-root node \(v\) has exactly one \emph{primary parent} (root or a previous node), which decides where its attempt begins: the discovery agent resumes the parent's saved workspace and inherits its accumulated observations as context to generate the attempt in \(v\). Node \(v\) keeps all previous generation history for it, and its own filesystem snapshot, together with an evaluation score \(s_v\). Scores are generated from a fixed protocol, and larger values are better. We define the root score $s_r = 0$. In both online rollout and offline evaluation, the exploration policy observes a tree \(\mathcal T\), initially
containing only the root, and selects nodes in \(\mathcal T\) to continue exploration. The eligible nodes are \(A(\mathcal T)=\{r\}\cup\{v\in\mathcal T: v \text{ is a leaf}\}\), which contains the root and all leave node in the current tree. We assume there are \(W\geq 1\) parallel workers, each running one generation--evaluation request at a time  (e.g. concurrent API calls). For the current tree \(\mathcal T\), the exploration policy’s action is a batch \(C\in A(\mathcal T;W)\), where \(A(\mathcal T;W)=\{C\subseteq A(\mathcal T): |C|\leq W\}\) is the feasible batch set. The batch determines both where exploration continues and how many attempts run concurrently. Both the online and offline phase use this
decision interface, and differ only in how state transitions after an action: one executes for real, the other replays.

\paragraph{Online rollout.} Let \(t=1,2,\ldots\) index iterations of online rollout, starting from policy \(\pi_1\) and empty history \(\mathcal H_0=()\). At iteration \(t\), the fixed policy \(\pi_t\) guides the discovery agent to generate a new online rollout with access to the
completed discovery history \(\mathcal H_{t-1}\) as context. The history informs exploration but is kept separate from the tree under construction.
% Let \(\mathcal T_{t}^k\) denote the new discovery tree after \(k\) completed decision rounds, with \(\mathcal T_{t}^0=\{r\}\). The rollout allows at most \(K_1\) rounds. At round $k \le K_1$, the \emph{exploration policy} chooses a node batch \(C_{t}^k \in A(\mathcal T_{t}^k;W)\) and each node \(v\in C_{t}^k\) is assigned to a worker. The discovery agent uses \(v\)'s saved workspace and available context to produce a new candidate, and the evaluator assesses the result. These attempts run in parallel, each producing one new child of its selected parent. Attaching the completed children to the current tree yields \(\mathcal T_{t}^{k+1}\), while all previously recorded nodes remain unchanged. This transition is stochastic because the discovery agent may generate different outcomes from the same starting workspace. For the next round, the newly created child becomes the selectable leaf of an extended branch, while the root remains selectable for opening further branches. The rollout ends when the policy selects an empty batch or completes \(K_1\) decision rounds. After the rollout terminates, its final tree is recorded as \(\mathcal T_t\) and appended to the history, giving \(\mathcal H_t= \mathcal H_{t-1} \cup \{\mathcal{T}_t\}\). The method then enters the offline phase using this expanded collection of replay worlds.
Let \(\mathcal T_t^k\) be the discovery tree after \(k\) completed rounds at iteration $t$, with \(\mathcal T_t^0=\{r\}\), and allow at most \(K_1\) rounds. At round \(k < K_1\), the policy picks \(C_t^k\in A(\mathcal T_t^k;W)\) and assigns each \(v\in C_t^k\) to a worker, which resumes \(v\)'s workspace, generates a candidate, and evaluates it. These attempts run in parallel and each adds one child to its selected parent. Thus, every \(v\in C_t^k\) is attached with a child, giving \(\mathcal T_t^{k+1}\). The transition from  \(\mathcal T_t^{k}\) to  \(\mathcal T_t^{k+1}\) is stochastic, since the agent may produce different outcomes from the same workspace. For the next round, each newly added node becomes the selectable leaf of its branch, while \(r\) stays selectable for opening new branches. The rollout ends when $C_t^k = \emptyset$ or after $K_1$ completed rounds. After the rollout terminates, its final tree is recorded as \(\mathcal T_t\) and appended to the history, giving \(\mathcal H_t= \mathcal H_{t-1} \cup \{\mathcal{T}_t\}\). The method then enters the offline evaluaton phase using \(\mathcal H_t\).

\paragraph{Offline evaluation.}
With $\mathcal H_t$ fixed, our method constructs and evaluates
$M\geq 1$ policy versions $\pi_t^0,\ldots,\pi_t^{M-1}$, where
$\pi_t^0=\pi_t$. Each version is evaluated on every historical tree
$\mathcal T_i$, $i=1,\ldots,t$, under budget-control parameters
$\rho\in\mathcal R:=\{0,0.25,0.5,0.75,1\}$, where $\rho$ controls the
policy's budget preference. The resulting feedback drives the next
policy version. We index policy versions by $m$, replay worlds by $i$,
budget preferences by $\rho$, and decision rounds within one replay by
$k$. The outer index $t$ is fixed and omitted below. For each $(m,i,\rho)$, replay starts from
$\mathcal T_{i,\rho}^{m,0}=\{r\}$, where
$\mathcal T_{i,\rho}^{m,k}\subseteq\mathcal T_i$ is the subtree revealed
after $k$ completed rounds. The recorded tree $\mathcal T_i$ never
changes; only the observed portion grows. At round $k<K_2$, the current
policy $\pi_t^m$, conditioned on $\rho$, selects
$C_{i,\rho}^{m,k}\in A(\mathcal T_{i,\rho}^{m,k};W)$ from the revealed
observations. Unlike online execution, offline replay returns recorded
children of the selected nodes deterministically rather than generating
new candidates. Thus,
$\mathcal T_{i,\rho}^{m,k+1}
=\mathcal T_{i,\rho}^{m,k}
\cup\bigcup_{v\in C_{i,\rho}^{m,k}}
\operatorname{Child}(v;\mathcal T_i,\mathcal T_{i,\rho}^{m,k})$,
where
$\operatorname{Child}(v;\mathcal T_i,\mathcal T_{i,\rho}^{m,k})$
contains unobserved children of $v$ in $\mathcal T_i$ given the current
revealed subtree. For $v\neq r$, it is $v$'s unique recorded child, if
one exists. For $v=r$, it is the earliest-created child of $r$ outside
$\mathcal T_{i,\rho}^{m,k}$, opening one previously unrevealed branch.
In either case,
$\operatorname{Child}(v;\mathcal T_i,\mathcal T_{i,\rho}^{m,k})
=\emptyset$ when no unobserved continuation remains. Newly revealed
nodes expose their stored observations before the policy makes its next
decision. Replay permits at most $K_2$ rounds, counting each nonempty batch as one
completed round, and stops when $C_{i,\rho}^{m,k}=\emptyset$, after
$K_2$ completed rounds, or when
$\mathcal T_{i,\rho}^{m,k}=\mathcal T_i$.
Let $k_{i,\rho}^{m,\star}\in\{0,\ldots,K_2\}$ denote the number of
completed rounds at termination. The final observed subtree is
$\mathcal T_{i,\rho}^{m,k_{i,\rho}^{m,\star}}\subseteq\mathcal T_i$. Replay thus measures how far a policy pursues each branch, how it groups
attempts into parallel batches, and when it opens a new branch or stops.
These choices vary across policies and budget preferences, but every
branch is traversed in its recorded parent--child order, and no outcome
outside $\mathcal T_i$ is produced.

\paragraph{Replay objective.}
The replay objective balances discovery quality and generation cost
across budget preferences while also favoring parallel execution.
For each replay world $\mathcal T_i$ used for evaluation, define
$B_i:=|\mathcal T_i|-1>0$ and
$S_i:=\max_{v\in\mathcal T_i\setminus\{r\}}s_v>0$, the total number of
evaluated non-root nodes and the best recorded score, respectively.
For each $\rho\in\mathcal R$, let
$n_{i,\rho}^m:=
|\mathcal T_{i,\rho}^{m,k_{i,\rho}^{m,\star}}|-1$
be the number of discovery-agent calls represented by the replay.
Let $s_{i,\rho}^m$ be the best score among the revealed non-root nodes,
i.e.,
$s_{i,\rho}^m:=
\max_{v\in
\mathcal T_{i,\rho}^{m,k_{i,\rho}^{m,\star}}\setminus\{r\}}s_v$
when $n_{i,\rho}^m>0$, and set $s_{i,\rho}^m:=0$ otherwise.
We normalize the attained quality as
$a_{i,\rho}^m:=
\operatorname{clip}(s_{i,\rho}^m/S_i,0,1)$. The budget sweep produces the quality--cost points
$\{(n_{i,\rho}^m,a_{i,\rho}^m):\rho\in\mathcal R\}$.
A point is dominated if another point uses no more calls and attains no
lower quality, with at least one strict improvement.
Let
$(\bar n_{i,j}^m,\bar a_{i,j}^m)_{j=1}^{L_i^m}$
be the distinct nondominated points ordered by increasing call count,
and set $\bar a_{i,0}^m:=0$.
The replay score is
\begin{equation}
    V_i^m
    :=
    \underbrace{
        \frac{1}{B_i}
        \sum_{j=1}^{L_i^m}
        \bigl(\bar a_{i,j}^m-\bar a_{i,j-1}^m\bigr)
        \bigl(B_i-\bar n_{i,j}^m\bigr)
    }_{\text{normalized hypervolume}}
    -
    \lambda
    \underbrace{
        \frac{1}{|\mathcal R|}
        \sum_{\substack{\rho\in\mathcal R\\ n_{i,\rho}^m>0}}
        \frac{k_{i,\rho}^{m,\star}}{n_{i,\rho}^m}
    }_{\text{parallelism penalty}}.
    \label{eq:dream-rsi-replay-score}
\end{equation}
Here, $\lambda$ controls the strength of the parallelism penalty.
The first term measures the normalized area dominated by the
quality--cost Pareto frontier, rewarding policies that attain higher
quality with fewer discovery-agent calls. The second term measures the
average number of completed sequential rounds per call across the budget
sweep, thereby favoring policies that execute more calls in parallel;
replays with zero calls contribute zero to this term.

\paragraph{Policy improvement and selection.} A policy version \(\pi_t^m\) is evaluated by its average replay score over $\mathcal{H}_t$, denoted by \(V^m=\frac{1}{t}\sum_{i=1}^{t}V_i^m\). The offline phase starts  \(\pi_t^0=\pi_t\). At each step \(m=0,\ldots,M-2\), a \emph{policy-development agent} inspects the replay trajectories and scores of \(\pi_t^m\), along with feedback from earlier revisions, to identify successful decisions and recurring failures. It then rewrites the executable policy code into \(\pi_t^{m+1}\), which is evaluated on the same \(t\) replay worlds. The next online policy is the best-scoring candidate, \(\pi_{t+1}=\pi_t^{m^\star}\) with \(m^\star\in\operatorname*{arg\,max}_{m \in \{0, \cdots, M-1\}}V^m\). Since the current rollout policy $\pi_t = \pi_t^0$ is included among the candidates,
$V^{m^\star}\ge V^0$, so $\pi_{t+1}$ is never worse than $\pi_t$ on the average
replay score over $H_t$.

\section{Experiments}
We evaluate \textsc{Dream-RSI} on algorithm engineering, kernel optimization, mathematical optimization, and harness optimization. We primarily use Gemini-3.1-Pro for scientific discovery experiments, with additional Lasso experiments using Gemini-3.7-Flash to assess applicability across backbones.  For harness optimization, we use Claude Opus 4.8 with Claude Code. On the scientific discovery tasks, our main controlled baseline is Recursive Fixed Exploration. Both methods use identical discovery agents, evaluators, initialization, per-round budget limits, and the same hand-designed starting policy. From the second round onward, Recursive Fixed Exploration keeps its policy fixed, while \textsc{Dream-RSI} updates it through replay-based dreaming over accumulated discovery history. For harness optimization, we focus on end-to-end performance and omit the fixed-policy comparison due to its substantial cost: an additional full search is estimated to cost over \$100,000. We measure discovery cost as the cumulative number of discovery-agent calls.

\subsection{Kernel Engineering}
\label{subsec:kerneleng}

The goal of \textbf{kernel engineering} is to discover high-performance kernel implementations while preserving numerical correctness. We use four representative KernelBench tasks~\citep{ouyang2025kernelbench}: \textbf{VGG16}, \textbf{LayerNorm}, \textbf{ConvDiv}, and \textbf{ConvMax}. Candidates are scored by execution performance, measured as inverse runtime ($1/\mathrm{ms}$), subject to correctness checks against the reference implementation. With Gemini-3.1 Pro as the coding agent, we compare \textsc{Dream-RSI} against Recursive Fixed Exploration under an identical evaluation protocol and initialization, and against the strong PaceEvolve baseline~\citep{yan2026pacevolve}, whose maximum discovery time we cap at two days for fairness.

\paragraph{Results.} Figure~\ref{fig:kernelbench-scaling} shows discovery trajectories as the number of generations increases. On VGG16 and LayerNorm, \textsc{Dream-RSI} matches final performance with $2.43\times$ and $1.79\times$ fewer generations. On ConvDiv and ConvMax, under comparable budgets, it achieves $2.09\times$ and $1.44\times$ higher performance. Adapting the exploration policy across recursive rounds thus improves both the efficiency and the effectiveness of long-horizon discovery, and \textsc{Dream-RSI} surpasses PACEvolve in kernel performance (Detailed PACEvolve scaling curves are provided in Appendix~\ref{app:pacevolve_scaling_curves}.). Appendix~\ref{app:discovered-convdiv} gives the discovered ConvDiv kernel. Rather than optimizing convolution in isolation, it restructures the operator chain around downstream pooling and reduction. Adjacent convolution outputs are max-pooled immediately, while width pooling and subsequent reductions run directly on register-resident values, avoiding materialization of convolution and pooled feature maps. Linear operations, such as scaling, averaging, and bias accumulation, are deferred to a lightweight final reduction.

\begin{figure}[t]
    \centering
\definecolor{oursblue}{HTML}{0072B2}
\definecolor{baselinegray}{HTML}{6B7280}
\definecolor{paceorange}{HTML}{D55E00}

\usepgfplotslibrary{groupplots}

\begin{tikzpicture}

\begin{groupplot}[
  group style={
    group size=2 by 2,
    horizontal sep=1.00cm,
    vertical sep=1.15cm
  },
  width=7.3cm,
  height=3.5cm,
  xmin=0,
  axis lines=left,
  axis line style={black!55,line width=0.65pt},
  xmajorgrids=false,
  ymajorgrids=true,
  grid style={black!8,line width=0.30pt},
  scaled ticks=false,
  tick align=outside,
  tick style={black!55,line width=0.50pt},
  tick label style={font=\scriptsize,text=black!80},
  title style={font=\small\bfseries,yshift=3pt},
  xlabel style={font=\small,yshift=2pt},
  ylabel style={font=\small},
  clip=true,
  clip marker paths=true,
  unbounded coords=discard,
  filter discard warning=false
]

% ============================================================
% VGG16
% ============================================================

\nextgroupplot[
  title={VGG16},
  xmax=1050,
  ymin=0.35,
  ymax=0.57,
  xtick={0,250,500,750,1000},
  xticklabels=\empty,
  ytick={0.35,0.40,0.45,0.50,0.55},
  ylabel={Performance ($1/\mathrm{ms}$)},
  legend to name=sharedlegend,
  legend columns=3,
  legend style={
    draw=none,
    font=\small,
    column sep=1.2em
  }
]

% Dream-RSI
\addplot[
  const plot,
  color=oursblue,
  line width=2.55pt,
  mark=*,
  mark size=1.35pt,
  mark options={fill=oursblue, draw=oursblue}
]
table[x=budget,y=score]
{Main_Text/raw_data/data/vgg16_ours.dat};
\addlegendentry{\textsc{Dream-RSI}}

% Recursive Fixed Exploration
\addplot[
  const plot,
  color=baselinegray,
  line width=1.75pt,
  densely dashed,
  mark=square*,
  mark size=1.00pt,
  mark options={fill=white, draw=baselinegray, line width=0.65pt}
]
table[x=budget,y=score]
{Main_Text/raw_data/data/vgg16_baseline.dat};
\addlegendentry{Recursive Fixed Exploration}

% PaceEvolve Pro legend
\addlegendimage{
  color=paceorange,
  densely dashed,
  line width=1.6pt
}
\addlegendentry{PaceEvolve}

% Dream-RSI annotation
\node[
  anchor=north west,
  font=\scriptsize\bfseries,
  text=oursblue!80!black,
  fill=white,
  draw=oursblue!20,
  rounded corners=1pt,
  line width=0.3pt,
  inner xsep=3pt,
  inner ysep=1.7pt
]
at (axis description cs:0.34,0.65)
{\makecell{
  $2.43\times$ fewer discovery-agent calls \\
  (comparable performance)
}};

% PaceEvolve Pro: off-scale reference
\draw[
  paceorange,
  densely dashed,
  line width=1.6pt
]
(axis description cs:0,0.06) --
(axis description cs:1,0.06);

\node[
  anchor=south east,
  font=\scriptsize\bfseries,
  text=paceorange,
  fill=white,
  inner xsep=2pt,
  inner ysep=0.8pt
]
at (axis description cs:0.98,0.075)
{0.2002};

% ============================================================
% LayerNorm
% ============================================================

\nextgroupplot[
  title={LayerNorm},
  xmax=1050,
  ymin=1.02,
  ymax=1.15,
  xtick={0,250,500,750,1000},
  xticklabels=\empty,
  ytick={1.00,1.05,1.10,1.15}
]

% Dream-RSI
\addplot[
  const plot,
  color=oursblue,
  line width=2.55pt,
  mark=*,
  mark size=1.35pt,
  mark options={fill=oursblue, draw=oursblue}
]
table[x=budget,y=score]
{Main_Text/raw_data/data/layernorm_ours.dat};

% Recursive Fixed Exploration
\addplot[
  const plot,
  color=baselinegray,
  line width=1.75pt,
  densely dashed,
  mark=square*,
  mark size=1.00pt,
  mark options={fill=white, draw=baselinegray, line width=0.65pt}
]
table[x=budget,y=score]
{Main_Text/raw_data/data/layernorm_baseline.dat};

% Dream-RSI annotation
\node[
  anchor=north west,
  font=\scriptsize\bfseries,
  text=oursblue!80!black,
  fill=white,
  draw=oursblue!20,
  rounded corners=1pt,
  line width=0.3pt,
  inner xsep=3pt,
  inner ysep=1.7pt
]
at (axis description cs:0.335,0.55)
{\makecell{
  $1.79\times$ fewer discovery-agent calls \\
  (comparable performance)
}};

% PaceEvolve Pro: off-scale reference
\draw[
  paceorange,
  densely dashed,
  line width=1.6pt
]
(axis description cs:0,0.06) --
(axis description cs:1,0.06);

\node[
  anchor=south east,
  font=\scriptsize\bfseries,
  text=paceorange,
  fill=white,
  inner xsep=2pt,
  inner ysep=0.8pt
]
at (axis description cs:0.98,0.075)
{0.8264};

% ============================================================
% ConvDiv
% ============================================================

\nextgroupplot[
  title={ConvDiv},
  xmax=1050,
  ymin=0.40,
  ymax=2.00,
  xtick={0,250,500,750,1000},
  ytick={0.4,0.8,1.2,1.6,2.0},
  xlabel={Number of Generations},
  ylabel={Performance ($1/\mathrm{ms}$)}
]

% Dream-RSI
\addplot[
  const plot,
  color=oursblue,
  line width=2.55pt,
  mark=*,
  mark size=1.35pt,
  mark options={fill=oursblue, draw=oursblue}
]
table[x=budget,y=score]
{Main_Text/raw_data/data/convdiv_ours.dat};

% Recursive Fixed Exploration
\addplot[
  const plot,
  color=baselinegray,
  line width=1.75pt,
  densely dashed,
  mark=square*,
  mark size=1.00pt,
  mark options={fill=white, draw=baselinegray, line width=0.65pt}
]
table[x=budget,y=score]
{Main_Text/raw_data/data/convdiv_baseline.dat};

% PaceEvolve Pro: actual score within y-range
\addplot[
  color=paceorange,
  densely dashed,
  line width=1.8pt,
  mark=none,
  forget plot
]
coordinates {
  (0,0.6452)
  (1050,0.6452)
};

\node[
  anchor=south east,
  font=\scriptsize\bfseries,
  text=paceorange,
  fill=white,
  inner xsep=2pt,
  inner ysep=0.8pt
]
at (axis cs:1020,0.665)
{0.6452};

% Dream-RSI annotation
\node[
  anchor=north west,
  font=\scriptsize\bfseries,
  text=oursblue!80!black,
  fill=white,
  draw=oursblue!20,
  rounded corners=1pt,
  line width=0.3pt,
  inner xsep=3pt,
  inner ysep=1.7pt
]
at (axis description cs:0.035,0.95)
{\makecell{
  $2.09\times$ higher score \\
  (Similar Budget)
}};

% ============================================================
% ConvMax
% ============================================================

\nextgroupplot[
  title={ConvMax},
  xmax=1050,
  ymin=0.22,
  ymax=0.45,
  xtick={0,250,500,750,1000},
  ytick={0.25,0.30,0.35,0.40,0.45},
  xlabel={Number of Generations}
]

% Dream-RSI
\addplot[
  const plot,
  color=oursblue,
  line width=2.55pt,
  mark=*,
  mark size=1.35pt,
  mark options={fill=oursblue, draw=oursblue}
]
table[x=budget,y=score]
{Main_Text/raw_data/data/convmax_ours.dat};

% Recursive Fixed Exploration
\addplot[
  const plot,
  color=baselinegray,
  line width=1.75pt,
  densely dashed,
  mark=square*,
  mark size=1.00pt,
  mark options={fill=white, draw=baselinegray, line width=0.65pt}
]
table[x=budget,y=score]
{Main_Text/raw_data/data/convmax_baseline.dat};

% Dream-RSI annotation
\node[
  anchor=north west,
  font=\scriptsize\bfseries,
  text=oursblue!80!black,
  fill=white,
  draw=oursblue!20,
  rounded corners=1pt,
  line width=0.3pt,
  inner xsep=3pt,
  inner ysep=1.7pt
]
at (axis description cs:0.035,0.95)
{\makecell{
  $1.44\times$ higher score \\
  (Similar Budget)
}};

% PaceEvolve Pro: off-scale reference
\draw[
  paceorange,
  densely dashed,
  line width=1.6pt
]
(axis description cs:0,0.2654) --
(axis description cs:1,0.2654);

\node[
  anchor=south east,
  font=\scriptsize\bfseries,
  text=paceorange,
  fill=white,
  inner xsep=2pt,
  inner ysep=0.8pt
]
at (axis description cs:0.98,0.075)
{0.2654};

\end{groupplot}

% ============================================================
% Shared legend
% ============================================================

\node[anchor=south]
at ([yshift=-0.15cm]current bounding box.north)
{\pgfplotslegendfromname{sharedlegend}};

\end{tikzpicture}
    % \vspace{-1.5em}
    \caption{
\textbf{GPU kernel engineering results.}
Discovery performance of \textsc{Dream-RSI} and Recursive Fixed Exploration as a function of the number of generations.
On VGG16 and LayerNorm, \textsc{Dream-RSI} reaches comparable performance with $2.43\times$ and $1.79\times$ fewer generations, respectively.
On ConvDiv and ConvMax, it achieves $2.09\times$ and $1.44\times$ higher performance under comparable discovery budgets.
Higher is better for all tasks.
}
    \label{fig:kernelbench-scaling}
\end{figure}

\begin{table}[t!]
\centering
\caption{
Performance comparison on mathematical discovery tasks.
Higher is better for \textbf{Sum Diff} and \textbf{Circle Packing}, while lower is better for \textbf{Auto Correlation}.
Best results are shown in \textbf{bold}.
}
% \vspace{-1em}
\label{tab:downstream_results}
\small
\setlength{\tabcolsep}{1pt}
\vspace{0.1in}
\resizebox{0.76\textwidth}{!}{\begin{tabular}{llccc}
\toprule
\textbf{Method}
& \textbf{LLM}
& \textbf{Sum Diff ($\uparrow$)}
& \textbf{Auto Correlation ($\downarrow$)}
& \textbf{Circle Packing ($\uparrow$)} \\
\midrule

AlphaEvolve
& Gemini-2.0 Pro + Flash 
& -- &  1.455700 &  2.635862 \\

AlphaEvolveV2
& Gemini-2.0 Pro + Flash 
& 1.121936 & -- & \textbf{2.635983} \\

OpenEvolve 
& - 
& -- &  1.460000 &  - \\

CodeEvolve 
& -
& -- & -- & 2.635980 \\

ShinkaEvolve 
& Mixed 
& -- &  1.457800 &  2.635982 \\

TTS-Discovery
& Qwen3-8B 
& -- & -- &  \textbf{2.635983} \\

ThetaEvolve
& Distilled-Qwen3-8B 
& -- & 1.493000 & \textbf{2.635983} \\

EvoX
& Gemini-3.0-Pro 
& -- & 1.458900 & 2.635900 \\

SimpleTES
& GPT-OSS-120B 
& 1.143975 & \bf 1.453675 & \textbf{2.635983} \\

\midrule
\textit{Our System} \\
Recursive Fixed Exploration
& Gemini-3.1-Pro 
& 1.144047 & 1.456001 & \textbf{2.635983} \\

\textsc{Dream-RSI}
& Gemini-3.1-Pro  
& \textbf{1.145427}
& 1.456375
& \textbf{2.635983} \\

\bottomrule
\end{tabular}}
\end{table}

\subsection{Mathematics Optimization}
\label{subsec:mathopt}

We further evaluate \textsc{Dream-RSI} on three mathematical discovery tasks spanning discrete combinatorial optimization, geometric optimization, and functional optimization: the \textbf{Sum--Difference Problem}, \textbf{Circle Packing}, and \textbf{Autocorrelation Inequalities}. Formal definitions of the three tasks are provided in Appendix ~\ref{sec:detailed_task_description}. We use Gemini-3.1 Pro via the Gemini CLI as the discovery agent for both methods for 10 rounds. For each task, the agent iteratively proposes and evaluates candidate constructions or optimization procedures according to the task-specific objective.

\paragraph{Results.}
Table~\ref{tab:downstream_results} summarizes the results.
\textsc{Dream-RSI} achieves a Sum--Difference score of $1.145427$, outperforming SimpleTES and Recursive Fixed Exploration.
On Circle Packing, it reaches $2.635983$, matching the strongest reported result among the compared methods.
For Autocorrelation, \textsc{Dream-RSI} obtains $1.456375$, remaining competitive with existing discovery systems.

% \usepgfplotslibrary{groupplots}
\pgfplotsset{compat=1.18}

% ============================================================
% Scaling data
% Metric: arithmetic mean runtime over six real downstream tasks
% Lower is better.
% ============================================================

% ------------------------------------------------------------
% Gemini-3.1-Pro: Fixed Exploration
% ------------------------------------------------------------
\pgfplotstableread{
compute avg iter
110 5267.0500 1
220 5365.4000 2
330 5295.2333 3
440 4691.0833 4
550 3587.0667 5
}\probaseline

% ------------------------------------------------------------
% Gemini-3.1-Pro: Dream-RSI
% Iter 1 shares the same initialization
% ------------------------------------------------------------
\pgfplotstableread{
compute avg iter
110 5267.0500 1
119 5349.1500 2
147 5368.9000 3
234 5705.4333 4
317 2931.0000 5
}\proours

% ------------------------------------------------------------
% Gemini-3.7-Flash: Fixed Exploration
% Iter 1--2 have no downstream evaluation
% ------------------------------------------------------------
\pgfplotstableread{
compute avg iter
640  2735.6 1
1280 2988.3 2
1920 3011.1333 3
2560 2909.1833 4
3200 2516.6833 5
}\flashbaseline

% ------------------------------------------------------------
% Gemini-3.7-Flash: Dream-RSI
% Iter 1 has no downstream evaluation
% ------------------------------------------------------------
\pgfplotstableread{
compute avg iter
640  2735.6 1
976  2394.9167 2
1230 2378.0500 3
1529 2327.7000 4
1879 2350.5833 5
}\flashours

% ============================================================
% Main figure
% ============================================================
\begin{figure*}[t!]
\centering
% ============================================================
% (a) Final performance table
% ============================================================

\resizebox{\textwidth}{!}{%
\begin{tabular}{l r r rr rrrr r}
\toprule

\textbf{Method}
& \textbf{Model}
& \textbf{Compute}
& \multicolumn{2}{c}{\textit{Non-biological}}
& \multicolumn{4}{c}{\textit{Biological}}
& \textbf{Avg.}
\\

\cmidrule(lr){4-5}
\cmidrule(lr){6-9}

&
&
&
\textbf{Gisette}
&
\textbf{RCV1}
&
\textbf{DNA}
&
\textbf{Leukemia}
&
\textbf{Colon}
&
\textbf{Duke Breast}
&
\\

\midrule

\multicolumn{10}{l}{\textit{Previous solvers}} \\[1pt]

sklearn
& --
& --
& 11275.2
& 252881.7
& 93.8
& 227.2
& 229.8
& 374.0
& 44180.3
\\

glmnet
& --
& --
& 9063.6
& 73072.8
& 351.9
& 45.0
& 24.2
& 47.7
& 13767.5
\\

SimpleTES
& gpt-oss-120b
& 51{,}200
& 3141.9
& 19625.6
& 15.9
& 15.5
& 11.6
& 18.1
& 3804.8
\\

SimpleTES $\dagger$
& gpt-oss-120b
& 51{,}200
& 8651.0
& 41143.1
& 37.6
& 28.2
& 19.5
& 31.1
& 8318.4
\\

\midrule

\multicolumn{10}{l}{\textit{Our System}} \\[1pt]

Recursive Fixed Exploration
& Gemini-3.1-Pro
& 550
& 1861.8
& 19550.1
& 41.5
& 26.1
& 14.5
& 28.4
& 3587.1
\\

&
Gemini-3.7-Flash
& 3200
& 1133.1
& 13873.0
& 29.8
& 24.1
& 15.7
& 24.4
& 2516.7
\\

\textbf{\textsc{Dream-RSI}}
& Gemini-3.1-Pro
& 317
& 2841.0
& 14616.0
& 49.9
& 30.2
& 16.4
& 32.5
& 2931.0
\\

&
Gemini-3.7-Flash
& 1879
& 1091.9
& 12923.4
& 31.4
& 21.0
& 12.2
& 23.6
& \textbf{2350.6}
\\

\bottomrule
\end{tabular}%
}

\vspace{0.08cm}

{\small\textbf{(a) Final performance of running time (ms).}}

\vspace{0.27cm}

% ============================================================
% (b) Scaling curves
% ============================================================

\begin{tikzpicture}

\begin{groupplot}[
group style={
  group size=2 by 1,
  horizontal sep=0.1\textwidth
},
width=0.5\textwidth,
height=3.5cm,
grid=major,
major grid style={
  gray!16,
  line width=0.3pt
},
axis line style={
  black!55,
  line width=0.65pt
},
tick style={
  black!55,
  line width=0.5pt
},
tick align=outside,
tick label style={
  font=\scriptsize,
  text=black!80
},
xlabel style={
  font=\small,
  yshift=1pt
},
ylabel style={
  font=\small
},
title style={
  font=\small\bfseries,
  yshift=2pt
},
scaled ticks=false,
clip=false
]

% ============================================================
% Left: Gemini-3.1-Pro
% ============================================================

\nextgroupplot[
title={Gemini-3.1-Pro},
xmin=50,
xmax=600,
ymin=2800,
ymax=6000,
xtick={100,300,500},
ytick={3000,4000,5000,6000},
xlabel={Discovery Compute},
ylabel={Avg. Runtime (ms)},
xticklabel style={
  /pgf/number format/fixed,
  /pgf/number format/precision=0
},
yticklabel style={
  /pgf/number format/fixed,
  /pgf/number format/precision=0,
  /pgf/number format/1000 sep={,}
},
legend to name=scalinglegend,
legend columns=2,
legend style={
  draw=none,
  font=\footnotesize,
  /tikz/every even column/.append style={
    column sep=1.4em
  }
}
]

% ------------------------------------------------------------
% Gemini-3.1-Pro: Fixed Exploration
% ------------------------------------------------------------

\addplot[
color=blue!75!black,
line width=1.7pt,
mark=square*,
mark size=2.5pt,
point meta=explicit symbolic,
nodes near coords,
every node near coord/.append style={
  font=\tiny,
  text=blue!75!black,
  anchor=north,
  yshift=-1.5pt,
  fill=white,
  fill opacity=0.90,
  text opacity=1,
  inner sep=0.7pt
}
]
table[
x=compute,
y=avg,
meta=iter
] {\probaseline};

\addlegendentry{Recursive Fixed Exploration}

% ------------------------------------------------------------
% Gemini-3.1-Pro: Dream-RSI
% ------------------------------------------------------------

\addplot[
color=red!80!black,
line width=1.9pt,
mark=*,
mark size=2.8pt,
point meta=explicit symbolic,
nodes near coords,
every node near coord/.append style={
  font=\tiny,
  text=red!80!black,
  anchor=south,
  yshift=1.5pt,
  fill=white,
  fill opacity=0.90,
  text opacity=1,
  inner sep=0.7pt
}
]
table[
x=compute,
y=avg,
meta=iter
] {\proours};

\addlegendentry{\textsc{Dream-RSI}}

% ============================================================
% Right: Gemini-3.7-Flash
% ============================================================

\nextgroupplot[
title={Gemini-3.7-Flash},
xmin=500,
xmax=3400,
ymin=2250,
ymax=3100,
xtick={1000,2000,3000},
ytick={2250,2500,2750,3000},
xlabel={Discovery Compute},
ylabel={},
xticklabel style={
  /pgf/number format/fixed,
  /pgf/number format/precision=0,
  /pgf/number format/1000 sep={,}
},
yticklabel style={
  /pgf/number format/fixed,
  /pgf/number format/precision=0,
  /pgf/number format/1000 sep={,}
}
]

% ------------------------------------------------------------
% Gemini-3.7-Flash: Fixed Exploration
% ------------------------------------------------------------

\addplot[
color=blue!75!black,
line width=1.7pt,
mark=square*,
mark size=2.5pt,
point meta=explicit symbolic,
nodes near coords,
every node near coord/.append style={
  font=\tiny,
  text=blue!75!black,
  anchor=south,
  yshift=1.5pt,
  fill=white,
  fill opacity=0.90,
  text opacity=1,
  inner sep=0.7pt
}
]
table[
x=compute,
y=avg,
meta=iter
] {\flashbaseline};

% ------------------------------------------------------------
% Gemini-3.7-Flash: Dream-RSI
% ------------------------------------------------------------

\addplot[
color=red!80!black,
line width=1.9pt,
mark=*,
mark size=2.8pt,
point meta=explicit symbolic,
nodes near coords,
every node near coord/.append style={
  font=\tiny,
  text=red!80!black,
  anchor=south,
  yshift=1.5pt,
  fill=white,
  fill opacity=0.90,
  text opacity=1,
  inner sep=0.7pt
}
]
table[
x=compute,
y=avg,
meta=iter
] {\flashours};

\end{groupplot}

\end{tikzpicture}

% ============================================================
% Shared legend
% ============================================================

\vspace{0.00cm}

{\centering
\pgfplotslegendfromname{scalinglegend}
\par
}

\vspace{0.015cm}

{\small\textbf{(b) Recursive Discovery Dynamics.}}

\vspace{-0.05cm}

% ============================================================
% Caption
% ============================================================

\caption{
% \textbf{Lasso regularization-path discovery results.}
% \textbf{(a)} Final wall-clock runtime on six held-out downstream tasks;
% lower is better.
% \textbf{Compute} denotes the cumulative number of discovery-agent calls.
% \textbf{(b)} \textbf{Recursive discovery dynamics.} Average downstream runtime across six held-out tasks versus cumulative discovery compute for Gemini-3.1-Pro and Gemini-3.7-Flash. Numbers next to markers denote recursive rounds (iterations). Lower is better.
\textbf{Lasso regularization-path discovery.}
\textbf{(a)} Final wall-clock runtime on six held-out tasks (lower is better). \textbf{Compute} is the cumulative number of discovery-agent calls.
\textbf{(b)} \textbf{Recursive discovery dynamics.} Average runtime on the same six tasks versus cumulative discovery compute for Gemini 3.1-Pro/3.7-Flash; marker labels give the round. $\dagger$ denotes results reproduced by us.
}

\label{fig:lasso-main}

\end{figure*}
\subsection{Algorithm Engineering}
\label{subsec:lasso}

In this task, we consider \textbf{Lasso Regularization Path} as our algorithm-engineering task, a fundamental computational primitive in high-dimensional statistics that is widely used in model selection and cross-validation across domains such as genomics and finance. We follow the benchmark setting of SimpleTES~\citep{ye2026evaluation}, where the goal is to discover efficient implementations of the complete Lasso regularization path while preserving numerical correctness. During discovery, we use the same 17 synthetic instances as SimpleTES, which cover diverse problem regimes in terms of dimensionality, sparsity, feature correlation, and active-set structure. To evaluate whether the discovered algorithms generalize beyond the search distribution, we additionally evaluate them on six held-out downstream datasets spanning biological and non-biological domains. We compare against standard Lasso solvers
\textbf{sklearn}~\citep{pedregosa2011scikit} and
\textbf{glmnet}~\citep{friedman2010regularization},
as well as \textbf{SimpleTES}~\citep{ye2026evaluation}. We run both controlled baseline and \textsc{Dream-RSI} for 5 rounds. 

\paragraph{Main Results.}
Figure~\ref{fig:lasso-main}(a) summarizes the Lasso discovery results.
Across both discovery-agent backbones, \textsc{Dream-RSI} achieves a better
downstream quality--compute trade-off than Recursive Fixed Exploration.
With Gemini-3.1 Pro, it reduces the average runtime across the six held-out
datasets from 3587.1\,ms to 2931.0\,ms while using only 317 discovery-agent
calls, compared with 550 calls for fixed exploration.
With Gemini-3.7-Flash, \textsc{Dream-RSI} further reduces the average runtime
from 2516.7\,ms to 2350.6\,ms using 1879 calls instead of 3200. Despite using substantially less discovery compute, the resulting solvers also
outperform the standard sklearn and glmnet implementations on all six held-out
datasets. Relative to SimpleTES's reported 51,200 generations, our Pro
and Flash runs use approximately 162$\times$ and 27$\times$
fewer discovery-agent calls, respectively, with lower average
downstream runtime, although these comparisons use different
model backbones. For 3.1 Pro, the average improvement over fixed exploration is
driven by RCV1, despite slower runtimes on the other five datasets.
For 3.7 Flash, runtime improves on five of the six datasets.

\paragraph{Recursive Discovery Dynamics.}
Figure~\ref{fig:lasso-main}(b) shows the trajectory of downstream performance across recursive discovery rounds against cumulative discovery compute. By design, both methods behave identically in the first round. Afterward, Recursive Fixed Exploration keeps a static exploration policy, while \textsc{Dream-RSI} iteratively refines and redeploys its policy by dreaming over accumulated discovery history. By the final round, \textsc{Dream-RSI} achieves lower average downstream runtime with fewer discovery-agent calls than Recursive Fixed Exploration on both backbones. We further examine the discovered solver, whose implementation appears in Appendix~\ref{app:discovered-lasso}. Unlike \textsc{SimpleTES}, which switches between LARS and coordinate descent by problem dimension, the discovered solver makes the active-set optimization itself adaptive. It pairs strong-rule screening with Cauchy--Schwarz KKT pruning, computing exact gradients only when necessary and falling back to a full refresh once pruning loses effectiveness.

\subsection{Harness Optimization}
\label{subsec:harnessopt}

\definecolor{dreamblue}{HTML}{0072B2}
\definecolor{baselinegray}{HTML}{6B7280}
\definecolor{ccgray}{HTML}{BFC3C9}
\definecolor{tkgray}{HTML}{9CA3AF}
\definecolor{mhgray}{HTML}{6B7280}

\begin{figure}[t!]
\centering
\vspace{-0.5em}
% ============================================================
% LEFT: (a)
% ============================================================
\begin{minipage}[t]{0.47\linewidth}
\centering
\vspace{0pt}

\begin{tikzpicture}
\begin{axis}[
    width=7.4cm,
    height=3.2cm,
    xmin=-1.5,
    xmax=66,
    ymin=0.715,
    ymax=0.818,
    axis lines=left,
    axis line style={black!55,line width=0.65pt},
    xmajorgrids=false,
    ymajorgrids=true,
    grid style={black!8,line width=0.30pt},
    scaled ticks=false,
    tick align=outside,
    tick style={black!55,line width=0.50pt},
    tick label style={font=\scriptsize,text=black!80},
    label style={font=\small},
    xlabel={Number of Generations},
    ylabel={TB2.1 Score},
    xtick={0,10,20,30,40,50,60},
    ytick={0.70,0.72,0.74,0.76,0.78,0.80},
    clip=true
]

% Dream-RSI best-so-far
\addplot[
    color=dreamblue,
    line width=2.55pt,
    mark=none
]
coordinates {
    (0,0.7247)
    (1,0.7247)
    (1,0.7753)
    (3,0.7753)
    (3,0.7865)
    (16,0.7865)
    (16,0.7921)
    (30,0.7921)
    (30,0.8034)
    (52,0.8034)
    (52,0.8090)
    (64,0.8090)
};

% Record-setting points
\addplot[
    only marks,
    mark=*,
    mark size=1.6pt,
    color=dreamblue
]
coordinates {
    (1,0.7753)
    (3,0.7865)
    (16,0.7921)
    (30,0.8034)
    (52,0.8090)
};

% Seed baseline
\addplot[
    color=baselinegray,
    densely dashed,
    line width=1.15pt
]
coordinates {
    (-1.5,0.7247)
    (66,0.7247)
};

\node[
    anchor=south east,
    font=\scriptsize,
    text=baselinegray,
    fill=white,
    inner xsep=2pt,
    inner ysep=0.6pt
]
at (axis cs:65,0.7262)
{Seed: 0.7247};

\node[
    anchor=north west,
    font=\scriptsize\bfseries,
    text=dreamblue!85!black,
    fill=white,
    draw=dreamblue!20,
    rounded corners=1pt,
    line width=0.3pt,
    inner xsep=3pt,
    inner ysep=1.7pt
]
at (axis description cs:0.26,0.62)
{$0.7247 \rightarrow 0.8090$};

\end{axis}
\end{tikzpicture}

\vspace{2pt}

{\small\textbf{(a)} Discovery Process}

\end{minipage}
\hfill
% ============================================================
% RIGHT: (b)
% ============================================================
\begin{minipage}[t]{0.42\linewidth}
\centering
\vspace{0pt}

\begin{tikzpicture}
\begin{axis}[
    width=6cm,
    height=3.0cm,
    ybar,
    bar width=15pt,
    xmin=0.5,
    xmax=4.5,
    ymin=0.68,
    ymax=0.82,
    xtick={1,2,3,4},
    xticklabels={CC,TK,MH, Dream-RSI},
    ytick={0.70,0.74,0.78,0.82},
    axis lines=left,
    axis line style={black!55,line width=0.65pt},
    xmajorgrids=false,
    ymajorgrids=true,
    grid style={black!8,line width=0.30pt},
    scaled ticks=false,
    tick align=outside,
    tick style={black!55,line width=0.45pt},
    tick label style={font=\scriptsize,text=black!80},
    label style={font=\small},
    xlabel={Method},
    ylabel={TB2.1 Score},
    nodes near coords,
    every node near coord/.append style={
        font=\scriptsize,
        yshift=1pt
    },
    clip=false
]

\addplot[
    draw=none,
    fill=ccgray,
    bar shift=0pt
]
coordinates {(1,0.753)};

\addplot[
    draw=none,
    fill=tkgray,
    bar shift=0pt
]
coordinates {(2,0.7247)};

\addplot[
    draw=none,
    fill=mhgray,
    bar shift=0pt
]
coordinates {(3,0.792)};

\addplot[
    draw=none,
    fill=dreamblue,
    bar shift=0pt
]
coordinates {(4,0.809)};

\end{axis}
\end{tikzpicture}

% \vspace{-1pt}

{\small\textbf{(b)} Detailed Comparison}

\end{minipage}
\vspace{-0.5em}
\caption{
Harness optimization results of \textsc{Dream-RSI}.
\textbf{(a)} Best-so-far TerminalBench2.1 score over the discovery process, improving the seed harness from 0.7247 to 0.8090.
\textbf{(b)} Full-suite comparison against Claude Code (CC), Terminus-Kira (TK), and Meta-Harness (MH).
}
\label{fig:dream-rsi-harness}

\end{figure}
We further evaluate \textsc{Dream-RSI} on the \emph{harness optimization} task, where the goal is to improve the program surrounding a fixed language model, including how information is stored, retrieved, and presented during agent execution. We follow the setting of Meta-Harness~\citep{lee2026meta} and initialize the discovery from Terminus-Kira as the seed harness. Each discovered harness is evaluated on the full Terminal-Bench 2.1~\citep{merrill2026terminal} suite of 89 tasks, with two independent runs per task. We use Claude Opus 4.8 with Claude Code as both the discovery agent and the policy-development agent, while harness evaluation is also performed with Claude Opus 4.8. We reproduce Meta-Harness following its original search protocol, using Opus 4.8 for both search and evaluation.

\paragraph{Results.} Figure~\ref{fig:dream-rsi-harness} shows that \textsc{Dream-RSI} improves the harness throughout the discovery process, increasing the full-suite score from 0.7247 to 0.8090. The discovered harness also outperforms Claude Code (0.75) and Meta-Harness (0.79) under the same evaluation setting.

\section{Further Analysis}
\paragraph{Analysis of Replay World Accumulation}
\begin{wraptable}[6]{r}{0.5\linewidth}
\vspace{-1em}
\centering
\caption{Effect of accumulated discovery history on \textsc{ConvDiv}.}
\label{tab:ablation-convdiv}
\vskip -0.1in
\small
\resizebox{0.47\textwidth}{!}{\begin{tabular}{@{}lcc@{}}
\toprule
\textbf{Method}
& \textbf{Performance} $\uparrow$
& \textbf{Compute} $\downarrow$ \\
\midrule
\textsc{Dream-RSI} (All History)
& \textbf{1.90} & 786 \\
Current History Only
& 1.59 & 430 \\
\bottomrule
\end{tabular}}
% \vspace{-0.8em}
\end{wraptable}
We study the effect of accumulating replay worlds across recursive rounds. 
% Table~\ref{tab:ablation-convdiv} shows \textsc{Dream-RSI} using the full discovery history reaches a performance of 1.90, compared with 1.59 when replay is restricted to the current round. Accumulating replay worlds may help learn a more general exploration policy, reducing overly conservative exploration induced by fitting to a single world.
Table~\ref{tab:ablation-convdiv} shows that \textsc{Dream-RSI} achieves a performance of 1.90 using the full discovery history, compared with 1.59 when replay is restricted to the current round. Accumulating replay worlds may help learn a more general exploration policy and avoid overly conservative exploration associated with fitting to a single world.

\paragraph{Analysis of Historical Inductive Biases in Long-Horizon Discovery}
\definecolor{oursblue}{HTML}{0072B2}
\definecolor{baselinegray}{HTML}{6B7280}
\definecolor{guidanceorange}{HTML}{D55E00}
\definecolor{guidancegreen}{HTML}{009E73}

\begin{wrapfigure}[10]{r}{0.5\linewidth}
\vspace{-3em}
\centering
\begin{tikzpicture}
\begin{axis}[
  width=\linewidth,
  height=3.5cm,
  xmin=0,
  xmax=1050,
  ymin=0.40,
  ymax=2.00,
  axis lines=left,
  axis line style={black!55,line width=0.65pt},
  xmajorgrids=false,
  ymajorgrids=true,
  grid style={black!8,line width=0.30pt},
  scaled ticks=false,
  tick align=outside,
  tick style={black!55,line width=0.50pt},
  tick label style={font=\scriptsize,text=black!80},
  xlabel style={font=\small,yshift=2pt},
  ylabel style={font=\small},
  clip=true,
  clip marker paths=true,
  xtick={0,250,500,750,1000},
  ytick={0.4,0.8,1.2,1.6,2.0},
  xlabel={Number of Generations},
  ylabel={Performance ($1/\mathrm{ms}$)},
  legend style={
    draw=none,
    font=\scriptsize,
    at={(0.43,1.03)},
    anchor=south,
    legend columns=2,
    row sep=1pt,
    column sep=0.8em
  }
]

% ============================================================
% Dream-RSI
% ============================================================
\addplot[
  const plot,
  color=oursblue,
  line width=2.0pt,
  mark=*,
  mark size=1.2pt,
  mark options={
    fill=oursblue,
    draw=oursblue
  }
]
table[x=budget,y=score]
{Main_Text/raw_data/data/convdiv_ours.dat};

\addlegendentry{\textsc{Dream-RSI}}

% ============================================================
% Recursive Fixed Exploration
% ============================================================
\addplot[
  const plot,
  color=baselinegray,
  line width=1.5pt,
  densely dashed,
  mark=square*,
  mark size=1.0pt,
  mark options={
    fill=white,
    draw=baselinegray,
    line width=0.65pt
  }
]
table[x=budget,y=score]
{Main_Text/raw_data/data/convdiv_baseline.dat};

\addlegendentry{Fixed Exploration}

% ============================================================
% Dream-RSI + History as Guidance
% ============================================================
\addplot[
  const plot,
  color=guidanceorange,
  line width=1.8pt,
  densely dashdotted,
  mark=triangle*,
  mark size=1.15pt,
  mark options={
    fill=guidanceorange,
    draw=guidanceorange
  }
]
table[x=budget,y=score]
{Main_Text/raw_data/data/convdiv_ours_guidance.dat};

\addlegendentry{\textsc{Dream-RSI} + Guidance}

% ============================================================
% Fixed Exploration + History as Guidance
% ============================================================
\addplot[
  const plot,
  color=guidancegreen,
  line width=1.6pt,
  densely dotted,
  mark=diamond*,
  mark size=1.05pt,
  mark options={
    fill=white,
    draw=guidancegreen,
    line width=0.65pt
  }
]
table[x=budget,y=score]
{Main_Text/raw_data/data/convdiv_baseline_guidance.dat};

\addlegendentry{Fixed + Guidance}

\end{axis}
\end{tikzpicture}

\vspace{-0.6em}
\caption{
ConvDiv discovery results. Replaying history beats simply using it as guidance.
}
\label{fig:convdiv_guidance}
\vspace{-0.8em}
\end{wrapfigure}

We further investigate how the nature of the historical inductive bias affects long-horizon discovery. A natural alternative for utilizing history is to abstract prior trajectories into high-level directional insights, which are directly injected into the prompt as explicit semantic guidance for subsequent rounds. To evaluate the efficacy of this prompt-level semantic guidance, we apply it to both Recursive Fixed Exploration and \textsc{Dream-RSI}. As illustrated in Figure~\ref{fig:convdiv_guidance}, explicit directional guidance yields lower final performance for both methods on ConvDiv. This suggests that such guidance may constrain exploration.

\paragraph{Analysis of Evolution of Exploration Behavior}
% ============================================================
% Preamble
% ============================================================
\pgfplotsset{compat=1.18}

\definecolor{oursblue}{HTML}{2563EB}
\definecolor{barblue}{HTML}{BFDBFE}

% ============================================================
% Data
% ============================================================

\pgfplotstableread{
round perf
0 0.427
1 0.625
2 0.855
3 1.403
4 1.488
5 1.499
6 1.770
7 1.880
8 1.898
}\perfdata

\pgfplotstableread{
round attempts
0 110
1 110
2 87
3 80
4 50
5 92
6 80
7 91
8 86
}\attemptdata

% ============================================================
% Figure
% ============================================================

\begin{wrapfigure}[17]{r}{0.48\textwidth}
% ------------------------------------------------------------
% (a) Round-best performance
% ------------------------------------------------------------
\begin{tikzpicture}
\begin{axis}[
    width=\linewidth,
    height=2.6cm,
    xmin=-0.4,
    xmax=8.4,
    ymin=0.35,
    ymax=2.03,
    xtick={0,1,2,3,4,5,6,7,8},
    xticklabels={E0,E1,E2,E3,E4,E5,E6,E7,E8},
    ylabel={Performance (1/ms)},
    axis lines=left,
    axis line style={black!55,line width=0.6pt},
    ymajorgrids=true,
    grid style={black!8,line width=0.3pt},
    tick align=outside,
    tick style={black!55},
    tick label style={font=\scriptsize},
    label style={font=\scriptsize},
    clip=false,
]

\addplot[
    color=oursblue,
    line width=1.5pt,
    mark=*,
    mark size=2.4pt,
    mark options={
        fill=white,
        draw=oursblue,
        line width=1.0pt
    }
]
table[x=round,y=perf] {\perfdata};

\addplot[
    only marks,
    mark=none,
    point meta=explicit symbolic,
    nodes near coords,
    every node near coord/.append style={
        font=\tiny\bfseries,
        text=oursblue,
        yshift=5pt
    }
]
table[
    x=round,
    y=perf,
    meta=perf
] {\perfdata};

\end{axis}
\end{tikzpicture}

\vspace{-2mm}

{\scriptsize\textbf{(a) Round-best performance}}

\vspace{1mm}

% ------------------------------------------------------------
% (b) Evaluated attempts
% ------------------------------------------------------------
\begin{tikzpicture}
\begin{axis}[
    width=\linewidth,
    height=2.6cm,
    xmin=-0.4,
    xmax=8.4,
    ymin=0,
    ymax=125,
    xtick={0,1,2,3,4,5,6,7,8},
    xticklabels={E0,E1,E2,E3,E4,E5,E6,E7,E8},
    ylabel={Evaluated attempts},
    xlabel={Recursive execution round},
    axis lines=left,
    axis line style={black!55,line width=0.6pt},
    ymajorgrids=true,
    grid style={black!8,line width=0.3pt},
    tick align=outside,
    tick style={black!55},
    tick label style={font=\scriptsize},
    label style={font=\scriptsize},
    clip=false,
]

% bars
\addplot[
    ybar,
    bar width=14pt,
    fill=barblue,
    draw=oursblue,
    line width=0.7pt
]
table[x=round,y=attempts] {\attemptdata};

% line connecting bar tops
\addplot[
    color=oursblue,
    line width=1.3pt,
    mark=*,
    mark size=2.2pt,
    mark options={
        fill=white,
        draw=oursblue,
        line width=1.0pt
    }
]
table[x=round,y=attempts] {\attemptdata};

% value labels
\addplot[
    only marks,
    mark=none,
    point meta=explicit symbolic,
    nodes near coords,
    every node near coord/.append style={
        font=\tiny\bfseries,
        text=oursblue,
        yshift=7pt
    }
]
table[
    x=round,
    y=attempts,
    meta=attempts
] {\attemptdata};

\end{axis}
\end{tikzpicture}

\vspace{-2mm}

{\scriptsize\textbf{(b) Exploration effort}}
\vspace{-0.5em}
\caption{
\textbf{Evolution of exploration behavior on ConvDiv.}
\textbf{(a)} Round-best performance across recursive execution rounds.
\textbf{(b)} The number of evaluated attempts in each round.
}
\label{fig:convdiv-dynamics}

\end{wrapfigure}
Figure~\ref{fig:convdiv-dynamics} illustrates how the learned exploration policy evolves across recursive rounds on ConvDiv. The exploration policy exhibits an adaptive pattern: as performance improves, it initially conserves discovery compute (e.g., reducing the number of evaluated attempts from 110 to 50). When progress subsequently plateaus, it increases exploration effort again, coinciding with further performance gains.

\section{Related Work}
\paragraph{AI-Driven Scientific and Algorithmic Discovery.}
LLM-based discovery systems iteratively generate, evaluate, and refine candidate solutions using prior artifacts and feedback, as in AlphaEvolve~\citep{novikov2025alphaevolve}, OpenEvolve~\citep{openevolve}, CodeEvolve~\citep{assumpccao2025codeevolve}, ShinkaEvolve~\citep{lange2026shinkaevolve}, PACEvolve~\citep{yan2026pacevolve}, DeltaEvolve~\citep{jiang2026deltaevolve} and MLEvolve~\cite{du2026mlevolve}. More recent work emphasizes the importance of exploration itself: SkyDiscover provides adaptive discovery infrastructure~\citep{liu2026skydiscover}, SwarmResearch dynamically orchestrates multiple search branches~\citep{virk2026swarmresearch}, and EvoX~\citep{liu2026evox} explicitly optimizes search strategies rather than only candidate solutions. This shift makes exploration a meta-level optimization problem, but useful supervision for exploration strategies is expensive and delayed because their quality often becomes apparent only after long discovery rollouts.

\paragraph{Self-Evolving Agents.}
A broader line of work studies agents that improve their own components during interaction. Prior methods evolve model weights~\citep{huang2026r,huang2026g,li2026mm}, agent harnesses~\citep{lee2026meta,zhang2026darwin}, contexts~\citep{zhang2026agentic}, skills~\citep{zhang2026coevoskills,ouyang2026skillos,wu2026agenticrectune}, model behavior through test-time learning~\citep{wang2025thetaevolve,yuksekgonul2026learning,yan2026pacevolve++,wu2026teaching}, rubrics~\citep{xiong2026rubrics}, environments~\citep{huang2026envharness,li2026recursive,li2026long} and other applications~\citep{dai2026takes}. Most operate at the \emph{object level}, improving components used for task execution or reasoning. Recent work has begun to optimize meta-level mechanisms, including search strategies and self-improvement procedures~\citep{liu2026evox,yan2026pacevolve++,wang2026metaskill,zhang2026hyperagents,kim2026meta}. However, such meta-level strategies are difficult to improve because their quality is often revealed only after costly long-horizon rollouts. \textsc{Dream-RSI} turns accumulated discovery history into replay simulators for off-policy evaluation and refinement of exploration policies, then redeploys the selected policy for new online discovery, closing a recursive self-improvement loop.

\paragraph{Memory, History, and Experience Reuse.}
Prior work reuses agent experience as search history, context, memory, reusable skills, or training signals. DeltaEvolve structures evolutionary history through semantic deltas~\citep{jiang2026deltaevolve}; SwarmResearch and MLEvolve use cross-branch or retrospective information to guide subsequent search~\citep{virk2026swarmresearch,du2026mlevolve}; and other work improves how agents access and retain experience through evolving contexts, broader harness state, libraries, or skills~\citep{zhang2026agentic,lee2026meta,xu2026test,ouyang2026skillos}. We take a different view: rather than using exploration history only as context or memory for the next decision, we organize it as a \emph{replay simulator} in which many alternative exploration controllers can be evaluated cheaply. This turns previously collected discovery experience into reusable feedback for meta-level optimization, alleviating the scarcity and high cost of training signals.

\section{Conclusion}
We presented \textsc{Dream-RSI}, a framework for the recursive self-improvement of exploration policies in long-horizon discovery. The central idea in \textsc{Dream-RSI} is to convert accumulated discovery history from static context into an active, replayable simulator, directly targeting the core bottleneck of meta-optimization: delayed and expensive feedback. By "dreaming" within replay simulators constructed from historical discovery trees, \textsc{Dream-RSI} evaluates candidate exploration policies rapidly and at negligible execution cost. Improved policies are then redeployed online, driving further discovery and expanding the simulator pool — closing the recursive self-improvement loop. Our experiments demonstrate competitive discovery quality across diverse domains and improved discovery-agent-call efficiency in several settings.

\newpage
\bibliography{main}

\appendix
\newpage
\section{AI Use Statement}
In this work, we used generative AI tools as the API in the experiments, as described in the paper: we use Gemini 3.1 Pro in the Section~\ref{subsec:kerneleng},\ref{subsec:mathopt},\ref{subsec:lasso}, Gemini 3.7 Flash in Section~\ref{subsec:lasso} and Claude Opus 4.8 with Claude Code in the Section\ref{subsec:harnessopt}. Additionally, we used Gemini 3.8 Flash for writing and grammar correction. We have reviewed all AI-assisted work. We take responsibility for the final content of this work, including text, claims or artifacts produced with the aid of generative AI.

\section{Detailed Experimental Settings}
\subsection{Evaluation Environments}

\paragraph{Lasso regularization-path discovery.}
We evaluate candidate solvers in a KVM virtual machine with an AMD EPYC 7B12 processor, 8 vCPUs, and 32\,GB RAM, running Debian-based Linux. The evaluation protocol follows \citet{ye2026evaluation}.

\paragraph{GPU kernel engineering.}
We evaluate candidate kernels on an NVIDIA L40S GPU. Numerical correctness is checked against the reference PyTorch implementation following the official KernelBench protocol~\citep{ouyang2025kernelbench}.

\paragraph{Mathematical optimization.}
We use the same hardware environment as for Lasso regularization-path discovery.

\paragraph{Harness optimization.}
We evaluate candidate harnesses on a CPU server with 96 physical cores (192 logical CPUs) and 708\,GiB RAM, using Docker to provide isolated task execution environments.

\subsection{Discovery Configuration}
\label{app:discovery-config}
Table~\ref{tab:discovery-config} summarizes the discovery configurations. Each offline phase evaluates the current policy and three newly generated policy versions ($M=4$). The parenthesized budget decomposition specifies the maximum number of branches and attempts per branch, including the initial attempt. These budgets are upper limits; actual discovery-agent calls depend on the policy's allocation and stopping decisions.
\begin{table}[h]
\centering
\caption{
Discovery configurations.
$T$ denotes the number of online discovery rounds,
$W$ the maximum number of parallel workers,
$B$ the maximum discovery-agent calls per online round,
and $M$ the number of candidate policies evaluated per offline phase,
including the current policy.
}
\label{tab:discovery-config}
\resizebox{0.7\linewidth}{!}{
\begin{tabular}{llrrrr}
\toprule
\textbf{Task} & \textbf{Discovery Model}
& $\boldsymbol{T}$ & $\boldsymbol{W}$
& $\boldsymbol{B}$ & $\boldsymbol{M}$ \\
\midrule
Lasso           & Gemini-3.1-Pro   & 5  & 10 & $110\;(10\times11)$ & 4 \\
Lasso           & Gemini-3.7-Flash & 5  & 32 & $640\;(32\times20)$ & 4 \\
\midrule
VGG16           & Gemini-3.1-Pro   & 9  & 10 & $110\;(10\times11)$ & 4 \\
LayerNorm       & Gemini-3.1-Pro   & 9  & 10 & $110\;(10\times11)$ & 4 \\
ConvDiv         & Gemini-3.1-Pro   & 9  & 10 & $110\;(10\times11)$ & 4 \\
ConvMax         & Gemini-3.1-Pro   & 9  & 10 & $110\;(10\times11)$ & 4 \\
\midrule
Sum--Difference & Gemini-3.1-Pro   & 10 & 10 & $110\;(10\times11)$ & 4 \\
Autocorrelation & Gemini-3.1-Pro   & 10 & 10 & $110\;(10\times11)$ & 4 \\
Circle Packing  & Gemini-3.1-Pro   & 10 & 10 & $110\;(10\times11)$ & 4 \\
\midrule
Harness Optimization        & Claude Opus 4.8  & 9  & 4  & $16\;(4\times4)$ & 4 \\
\bottomrule
\end{tabular}
}
\end{table}

\subsection{Detailed Task Description}
\label{sec:detailed_task_description}
\begin{taskbox}{Problem 1 (Lasso Regularization Path)}
% \small
Given a feature matrix
$X \in \mathbb{R}^{n \times p}$,
response $y \in \mathbb{R}^{n}$,
and a decreasing sequence
$\lambda_1 > \cdots > \lambda_K$,
define
\[
F_k(w)
:=
\frac{1}{2n}\|y-Xw\|_2^2
+
\lambda_k\|w\|_1,
\qquad
w_k^\star
\in
\arg\min_{w\in\mathbb{R}^p} F_k(w).
\]

A candidate solver returns approximate coefficients
$\widetilde{W}
=
(\widetilde{w}_1,\ldots,\widetilde{w}_K)$.
It passes the benchmark's objective-value check if
\[
F_k(\widetilde{w}_k)
\le
F_k(w_{k,\mathrm{sklearn}})
+
10^{-6},
\qquad
\text{for every } k,
\]
where correctness is checked on fresh instances distinct from the timing
instances. If any required check fails, the search score is zero.

Otherwise, letting $\mathcal{I}$ denote the timing instances and $t_i$ the
time required to compute the complete regularization path on instance $i$,
the search score is
\[
R_{\mathrm{search}}
=
\left(
\prod_{i\in\mathcal{I}} t_i
\right)^{-1/|\mathcal{I}|}.
\]

\end{taskbox}

\begin{taskbox}{Problem 2 (Sum--Difference Problem)}
\small

The Sum--Difference Problem asks for a finite set
$A \subset \mathbb{Z}$
whose normalized sumset is large relative to its normalized difference set.
The objective is
\[
\Gamma(A)
:=
\frac{
\log\!\left(|A+A|/|A|\right)
}{
\log\!\left(|A-A|/|A|\right)
},
\]
where
\[
A+A
:=
\{a+a' : a,a' \in A\},
\qquad
A-A
:=
\{a-a' : a,a' \in A\}.
\]

The goal is to construct a finite set $A \subset \mathbb{Z}$ that maximizes
$\Gamma(A)$.

\end{taskbox}

\begin{taskbox}{Problem 3 (Circle Packing in a Unit Square)}
\small

For $n \in \{26,32\}$, the circle-packing task asks for centers
$(x_i,y_i)\in[0,1]^2$
and radii $r_i\ge 0$ such that every circle lies inside the unit square
and no two circles overlap:
\[
r_i \le x_i \le 1-r_i,
\qquad
r_i \le y_i \le 1-r_i,
\]
\[
(x_i-x_j)^2+(y_i-y_j)^2
\ge
(r_i+r_j)^2,
\qquad
1\le i<j\le n.
\]

The objective is to maximize
\[
\sum_i r_i.
\]

\end{taskbox}

\begin{taskbox}{Problem 4 (Autocorrelation Inequalities)}
\small

For an integrable function $f:\mathbb{R}\to\mathbb{R}$, define the
autoconvolution
\[
(f*f)(t)
:=
\int_{\mathbb{R}} f(t-x)f(x)\,dx,
\qquad
t\in\left[-\frac12,\frac12\right].
\]

\textbf{First Autocorrelation Inequality.}
Find a non-negative integrable function
$f:\mathbb{R}\to\mathbb{R}$
supported on
$\left[-\frac14,\frac14\right]$
such that
\[
\int_{-1/4}^{1/4} f(x)\,dx = 1.
\]
The objective is to minimize
\[
\Phi_1(f)
:=
\max_{t\in[-1/2,\,1/2]} (f*f)(t).
\]

\textbf{Second Autocorrelation Inequality.}
Find a non-negative integrable function
$f:\mathbb{R}\to\mathbb{R}$
supported on
$\left[-\frac14,\frac14\right]$
such that
\[
\int_{-1/4}^{1/4} f(x)\,dx = 1.
\]
The objective is to maximize
\[
\Phi_2(f)
:=
\frac{\|f*f\|_2^2}
{\|f*f\|_1\|f*f\|_\infty}.
\]

\textbf{Third Autocorrelation Inequality.}
Find an integrable (possibly signed) function
$f:\mathbb{R}\to\mathbb{R}$
supported on
$\left[-\frac14,\frac14\right]$
such that
\[
\int_{-1/4}^{1/4} f(x)\,dx = 1.
\]
The objective is to minimize
\[
\Phi_3(f)
:=
\max_{t\in[-1/2,\,1/2]}
\left|(f*f)(t)\right|.
\]

\end{taskbox}

\begin{taskbox}{Problem 5 (VGG16)}
\small

The VGG16 task asks for an efficient implementation of the forward pass of a
VGG16 network. The input is
\[
x \in \mathbb{R}^{10\times 3\times 224\times 224},
\]
and the network consists of five convolutional blocks. Each convolution uses a
$3\times3$ kernel with stride $1$ and padding $1$, followed by a ReLU
activation. The channel configurations of the five blocks are
\[
(64,64),\quad
(128,128),\quad
(256,256,256),\quad
(512,512,512),\quad
(512,512,512),
\]
with a $2\times2$ max-pooling operation of stride $2$ after each block.

After the final pooling layer, the resulting
$512\times7\times7$ feature map is flattened and passed through three fully
connected layers,
\[
512\cdot7\cdot7
\;\rightarrow\;
4096
\;\rightarrow\;
4096
\;\rightarrow\;
1000,
\]
with ReLU activations after the first two layers. Dropout is disabled
($p=0$). The final output is
\[
y \in \mathbb{R}^{10\times1000}.
\]

The goal is to produce an implementation numerically equivalent to the
reference PyTorch computation while minimizing GPU execution time.

\end{taskbox}

\begin{taskbox}{Problem 6 (ConvDiv)}
\small

The ConvDiv task applies a sequence of 3D convolution, scaling, pooling, and
reduction operations. The input is
\[
x\in
\mathbb{R}^{128\times8\times16\times64\times64}.
\]

First, a 3D convolution with $8$ input channels, $16$ output channels, and a
$3\times3\times3$ kernel is applied without padding:
\[
z = \operatorname{Conv3D}(x)
\in
\mathbb{R}^{128\times16\times14\times62\times62}.
\]
The convolution output is divided by $2$ and then passed through
$2\times2\times2$ max pooling:
\[
u
=
\operatorname{MaxPool}_{2\times2\times2}
\left(\frac{z}{2}\right)
\in
\mathbb{R}^{128\times16\times7\times31\times31}.
\]

Global average pooling is subsequently applied over the three spatial
dimensions,
\[
v_{b,c}
=
\frac{1}{7\cdot31\cdot31}
\sum_{d,h,w} u_{b,c,d,h,w},
\]
giving
\[
v\in\mathbb{R}^{128\times16\times1\times1\times1}.
\]
A learnable channel-wise bias
\[
\beta\in\mathbb{R}^{16\times1\times1\times1}
\]
is then added, followed by summation over the channel dimension:
\[
y_b
=
\sum_{c=1}^{16}
\left(v_{b,c}+\beta_c\right).
\]
The output has shape
\[
y\in\mathbb{R}^{128\times1\times1\times1}.
\]

The goal is to compute this operator chain exactly up to the benchmark's
numerical tolerance while minimizing GPU execution time.

\end{taskbox}

\begin{taskbox}{Problem 7 (ConvMax)}
\small

The ConvMax task consists of a 3D convolution followed by max pooling,
channel-wise log-sum-exp reduction, and ReLU. The input is
\[
x\in
\mathbb{R}^{4\times32\times32\times128\times128}.
\]

A 3D convolution maps $32$ input channels to $64$ output channels using a
$3\times3\times3$ kernel with stride $1$ and padding $1$:
\[
z
=
\operatorname{Conv3D}(x)
\in
\mathbb{R}^{4\times64\times32\times128\times128}.
\]
This is followed by $2\times2\times2$ max pooling with stride $2$,
\[
u
=
\operatorname{MaxPool}_{2\times2\times2}(z)
\in
\mathbb{R}^{4\times64\times16\times64\times64}.
\]

The channel dimension is then reduced using log-sum-exp:
\[
s_{b,1,d,h,w}
=
\log
\left(
\sum_{c=1}^{64}
\exp(u_{b,c,d,h,w})
\right),
\]
and the final output is
\[
y
=
\operatorname{ReLU}(s)
\in
\mathbb{R}^{4\times1\times16\times64\times64}.
\]

The goal is to produce an implementation numerically equivalent to the
reference computation while minimizing GPU execution time.

\end{taskbox}

\begin{taskbox}{Problem 8 (Layer Normalization)}
\small

The LayerNorm task applies layer normalization to an input tensor
\[
x
\in
\mathbb{R}^{16\times64\times256\times256}.
\]
Normalization is performed independently for each batch element over the
remaining three dimensions. Let
\[
M = 64\cdot256\cdot256.
\]
For each batch element $b$, define
\[
\mu_b
=
\frac{1}{M}
\sum_{c,h,w} x_{b,c,h,w},
\]
and
\[
\sigma_b^2
=
\frac{1}{M}
\sum_{c,h,w}
\left(x_{b,c,h,w}-\mu_b\right)^2.
\]

The normalized output is
\[
y_{b,c,h,w}
=
\gamma_{c,h,w}
\frac{x_{b,c,h,w}-\mu_b}
{\sqrt{\sigma_b^2+\epsilon}}
+
\beta_{c,h,w},
\]
where
\[
\gamma,\beta
\in
\mathbb{R}^{64\times256\times256}
\]
are learnable affine parameters and
$\epsilon=10^{-5}$.

The output has the same shape as the input,
\[
y
\in
\mathbb{R}^{16\times64\times256\times256}.
\]

The goal is to match the reference LayerNorm computation within the
benchmark's numerical tolerance while minimizing GPU execution time.

\end{taskbox}

% \section{Additional Results}
\section{Results of PACEvolve~\citep{yan2026pacevolve} on KernelBench}
\label{app:pacevolve_scaling_curves}
\begin{figure}[t!]
    \centering
    % PACEvolve evolve curves -- 48h (2-day) wall-clock budget cap on every run.
% Data: plots/pacevolve/Main_Text/raw_data/data/{task}_pacevolve_pro.dat
% x = PACEvolve generation index, y = best-so-far performance (1/ms)
% Requires: \usepackage{pgfplots,makecell}

\definecolor{oursblue}{HTML}{0072B2}
\definecolor{baselinegray}{HTML}{6B7280}

\usepgfplotslibrary{groupplots}

\begin{tikzpicture}

\begin{groupplot}[
  group style={
    group size=2 by 2,
    horizontal sep=1.00cm,
    vertical sep=1.45cm
  },
  width=7.3cm,
  height=4.4cm,
  xmin=0,
  axis lines=left,
  axis line style={black!55,line width=0.65pt},
  xmajorgrids=false,
  ymajorgrids=true,
  grid style={black!8,line width=0.30pt},
  scaled ticks=false,
  tick align=outside,
  tick style={black!55,line width=0.50pt},
  tick label style={font=\scriptsize,text=black!80},
  title style={font=\small\bfseries,yshift=3pt},
  xlabel style={font=\small,yshift=2pt},
  ylabel style={font=\small},
  clip=true,
  clip marker paths=true,
  restrict x to domain=0:2000,
  unbounded coords=discard,
  filter discard warning=false
]

% ============================================================
% VGG16
% ============================================================

\nextgroupplot[
  title={VGG16},
  xmax=450,
  ymin=0.160,
  ymax=0.212,
  xtick={0,100,200,300,400},
  xticklabels=\empty,
  ytick={0.16,0.17,0.18,0.19,0.20,0.21},
  ylabel={Performance ($1/\mathrm{ms}$)}
]

\addplot[
  const plot,
  color=oursblue,
  line width=2.20pt,
  mark=*,
  mark size=0.85pt,
  mark options={
    fill=oursblue,
    draw=oursblue
  }
]
table[x=budget,y=score]
{Main_Text/raw_data/data/vgg16_pacevolve_pro.dat};

% ============================================================
% LayerNorm
% ============================================================

\nextgroupplot[
  title={LayerNorm},
  xmax=450,
  ymin=0.60,
  ymax=1.12,
  xtick={0,100,200,300,400},
  xticklabels=\empty,
  ytick={0.6,0.7,0.8,0.9,1.0,1.1}
]

\addplot[
  const plot,
  color=oursblue,
  line width=2.20pt,
  mark=*,
  mark size=0.85pt,
  mark options={
    fill=oursblue,
    draw=oursblue
  }
]
table[x=budget,y=score]
{Main_Text/raw_data/data/layernorm_pacevolve_pro.dat};

% ============================================================
% ConvDiv
% ============================================================

\nextgroupplot[
  title={ConvDiv},
  xmax=450,
  ymin=0.15,
  ymax=0.70,
  xtick={0,100,200,300,400},
  ytick={0.2,0.3,0.4,0.5,0.6,0.7},
  xlabel={Number of Generations},
  ylabel={Performance ($1/\mathrm{ms}$)}
]

\addplot[
  const plot,
  color=oursblue,
  line width=2.20pt,
  mark=*,
  mark size=0.85pt,
  mark options={
    fill=oursblue,
    draw=oursblue
  }
]
table[x=budget,y=score]
{Main_Text/raw_data/data/convdiv_pacevolve_pro.dat};

% ============================================================
% ConvMax
% ============================================================

\nextgroupplot[
  title={ConvMax},
  xmax=450,
  ymin=0.16,
  ymax=0.285,
  xtick={0,100,200,300,400},
  ytick={0.16,0.20,0.24,0.28},
  xlabel={Number of Generations}
]

\addplot[
  const plot,
  color=oursblue,
  line width=2.20pt,
  mark=*,
  mark size=0.85pt,
  mark options={
    fill=oursblue,
    draw=oursblue
  }
]
table[x=budget,y=score]
{Main_Text/raw_data/data/convmax_pacevolve_pro.dat};

\end{groupplot}

\end{tikzpicture}
    \caption{Evolution of the best-so-far kernel performance discovered by PACEvolve on four KernelBench tasks under a 48-hour wall-clock budget. Performance is measured as inverse runtime ($1/\mathrm{ms}$), with higher values indicating faster kernels.}
    \label{fig:pacevolve_kernelbench}
\end{figure}

We present the detailed discovery process of PACEvolve~\citep{yan2026pacevolve} on KernelBench in Figure \ref{fig:pacevolve_kernelbench}. For a controlled comparison, we allow PACEvolve a 48-hour wall-clock budget on each task and report its best-so-far performance throughout the search.

\newpage
\section{Prompts}
\label{app:prompts}

For reproducibility, we provide the prompts used for online exploration and
replay-based exploration-policy improvement.
Variables enclosed by dollar signs or braces are instantiated by the runtime
system before execution.

\subsection{Exploration Prompt}
\label{app:exploration-prompt}

The following prompt is used to guide the discovery agent during online
exploration. It requires the agent to inspect the complete available discovery
history before proposing a new solution, explicitly reason about both successful
and failed attempts, and avoid repeatedly exploiting a locally saturated
direction.

\begin{lstlisting}[
    style=promptstyle,
    caption={Prompt used for online exploration.},
    label={lst:exploration-prompt}
]
You must read every historical proposal before proposing or implementing a new solution.

$direction_guidance

Variables (`$node_dir`, `$history_dir`, `$baseline_dir`, `$eval_program`, `$problem_file`) are filled in by the calling system. `$node_dir` is your own attempt directory — exclude it when scanning sibling `attempt_*/` dirs.

## 1. Read the complete history first

Before proposing anything, read every `proposal.md` under sibling `attempt_*/` dirs, `$history_dir`, and `$baseline_dir` in full — not a sample, not just recent cycles or the current branch. For each, read its matching `eval/score.json` (and `error.txt` if it failed). Trust the measured result over what the proposal claims about itself.

## 2. Learn from both successes and failures

For every past attempt, note the mechanism and how it did. For failures, figure out *why*: a flawed core idea, or a good idea let down by a bug, bad parameters, or an implementation slip? Don't repeat the former. The latter is worth retrying — but only once you've actually located the bug in the code (not just guessed from the proposal), and only with a specific fix in hand.

## 3. Don't converge into a local optimum

Look at the shape of what's been tried. If most attempts cluster around small variations of one mechanism with flattening returns, that's a local optimum - resist proposing another small tweak there. Deliberately favor a structurally different mechanism or an untried combination over a safer marginal refinement. Exploration diversity matters as much as the next incremental gain.

## 4. Propose and implement

The new idea must be a genuinely new mechanism, a new combination of previously-successful pieces, or a targeted fix to a specific bug found in step 2 - never a repeat or rename of something already tried. Implement it in `$eval_program`. Don't claim it compiles, is correct, or beats SOTA until it's actually evaluated.

## Files

Write only `$node_dir/proposal.md` (mechanism, evidence from history, why it's not a repeat, expected benefit/risk) and `$node_dir/$eval_program`. Everything else is read-only.

## Note:
    Never execute pkill, kill, killall, or terminate unrelated processes.
\end{lstlisting}

\subsection{Replay-Based Policy Improvement Prompt}
\label{app:replay-prompt}

The following prompt is used by the controller-development agent during
historical replay. The agent modifies the exploration policy using feedback
obtained from replay over previously collected discovery trajectories while
remaining restricted to prefix-observable information.

\begin{lstlisting}[
    style=promptstyle,
    caption={Prompt used for replay-based improvement of the exploration policy.},
    label={lst:replay-prompt}
]
You are improving one **prefix-only exploration policy**. Edit only
``{method_file}`` and implement ``OptimalPolicy.solve(self, question, budget=None)``.
Do not solve the scientific task and do not edit any other program.

## Objective: quality, work, and parallelism

The environment is a frozen, irregular branch×attempt grid. A policy opens a root
or refines the next cell of an already-open branch. Each revealed cell costs one
probe. The policy sees only the cells it has revealed so far; unrevealed scores are
unknown.

The evaluator sweeps your single ``beta`` knob and ranks the resulting curve by:

    pareto.reward = pareto.auc - lambda * parallel_penalty

``pareto.auc`` rewards reaching high per-trace attainment with few **total probes**.
``parallel_penalty`` is the mean of
``effective_sequential_rounds / total_probes`` over the sweep. For a batch of size
``k`` with ``W = question.max_parallelism`` workers, it costs one decision round and
``ceil(k / W)`` effective sequential rounds. A serial policy has penalty near 1;
useful full batches approach ``1/W``. Therefore choose only promising probes, but
batch independent promising probes whenever possible.

A local implementation failure does not by itself prove that its parent direction
is poor. Weigh recovery value against new roots and ordinary refinements while
keeping batches parallel.

## API

    question.reset()
    question.observed() -> dict[str, Observation]   # revealed prefix only
    question.legal_actions() -> list[str]           # roots + opened-branch frontiers
    question.legal_roots() -> list[str]             # unopened roots only
    question.opened_branches() -> list[int]
    question.meta(cell_id) -> CellMeta              # .branch .attempt .parent_id .seq .tags
    question.probe_batch(cells, on_reveal=...) -> list[Observation]
    question.baseline_score
    question.max_parallelism

``Observation`` supplies ``branch``, ``attempt``, ``score``, ``evaluated``, ``valid``,
``fail_class``, ``error``, ``delta_vs_baseline``, ``delta_vs_parent``, ``n_valid``, and
``n_total``.
Use the helpers in ``see.policy.observation_signal`` when useful:
``branch_promising``, ``branch_failed_hard``, ``probe_improved_vs_parent``, and
``probe_improved_vs_baseline``.

**Success semantics:** an evaluated observation with ``error is None`` and
``fail_class == "ok"`` is a successful evaluation, even when ``valid == False`` or
``n_valid``/``n_total`` are unavailable. Never label it repairable solely because
``valid`` is false. A *successful anchor* below means the best historical score
from such a successful evaluation.

Do **not** use ``question.best_so_far`` or ``question.budget_spent`` to decide what
to explore; they are bookkeeping only. Derive any decision statistic from
``question.observed()`` instead.

## Required branch trajectory and failure interpretation

For each opened branch, reconstruct its ordered prefix trajectory, not only its
latest observation or best score: successful anchor, score trend, regressions,
failure/repair sequence, and explored versus remaining depth.

Before closing or deprioritizing a failed frontier, classify it as
hard-unrecoverable, repairable implementation failure, weak-but-underexplored, or
repeatedly unpromising after sufficient valid evidence. Output/correctness mismatch,
shared-memory/resource limits, and variable/code, mask/layout/shape errors are
normally repairable. Do not infer algorithmic failure from one such error.
``n_valid == 0`` and ``branch_failed_hard(obs)`` are signals, not unconditional
closure: use ``fail_class`` and ``error`` to distinguish a repairable zero-valid
failure from an environment/dependency failure. ``compile_other`` alone is not
permanently hard. Classify the current failure episode: a later successful result
reopens the branch and cancels closure based only on an earlier failure.

## Required batch decision loop

At each decision round:

1. Read the prefix, reconstruct trajectories, and close only branches with
   cumulative evidence of being hard-unrecoverable or repeatedly unpromising.
2. Rank legal roots and legal branch frontiers using only prefix-derived signals:
   successful anchor, parent→child gain, complete branch trajectory, actual success
   versus failure evidence,
   failure recoverability, prior repair outcomes, remaining depth, and cross-branch
   comparison.
3. Rank actual repairable failures and underexplored frontiers in deterministic
   queues using trajectory, recoverability, remaining depth, repeated failures, and
   beta. A repairable failure retains eligibility unless cumulative evidence lowers
   its relative priority.
4. Build one **dynamic portfolio** batch of independent candidates, up to
   ``question.max_parallelism``: exploitation (strong normal refinements),
   exploration (new roots or underexplored branches), and at most one recovery
   (an actual repairable failure). When multiple roles are eligible, give
   exploration and justified recovery representation before filling remaining slots
   by priority; adapt this to prefix evidence rather than fixed quotas. Recovery
   must not displace normal successful refinements or leave workers idle. Never
   sample randomly, and do not default to a singleton merely because its top
   candidate is clear.
5. Stop only after considering the whole revealed portfolio: active, underexplored,
   recoverable, unopened, and remaining legal candidates. Do not stop while an
   eligible high-priority recovery or underexplored candidate remains; every
   remaining action needs an evidence-based decision to continue, reserve, or close.

A batch must contain distinct cells that are all legal *before* the call. It may
contain several roots and/or one frontier from each opened branch. It must never
contain a parent and its child together. Do not use a fixed widen-all / deepen-all
wave schedule: adapt batch composition after every revealed prefix.

Minimal structure:

    from see.policy.api import (
        LLMDesignedMethod, SimResult, _budget_done, _record_curve, finalize_result,
    )

    def solve(self, question, budget=None):
        question.reset()
        res, closed = SimResult(), set()
        while not _budget_done(question, budget):
            prefix = question.observed()
            update_closed(closed, prefix, question)
            batch = select_batch(prefix, question, closed)
            if not batch:
                break
            question.probe_batch(
                batch,
                on_reveal=lambda _: _record_curve(res, question),
            )
        return finalize_result(question, res)

## Hard constraints

- Keep ``NAME = "OptimalPolicy"`` and implement
  ``class OptimalPolicy(LLMDesignedMethod)`` in ``{method_file}`` only.
- **Prefix-only:** decisions may use revealed observations, ``baseline_score``, legal
  sets, structural ``meta``, and helper signals. Never use unrevealed scores, a true
  optimum, hardcoded winning cell ids, absolute score targets, or internal trace data.
- Every prune, widen, deepen, batch, and stop decision must be explainable from the
  current prefix. Shallow weak scores are not enough to discard a branch: deeper
  attempts can recover. A repairable latest failure must not erase its historical
  successful anchor or by itself cause permanent starvation.
- Replay calls with ``budget=None``. Always terminate when no batch is selected; do
  not assume a budget cap exists.
- A selected batch must be legal, have no duplicate ids, and contain at most
  ``question.max_parallelism`` cells.

## Beta: fixed per run, adaptive across cycles

Read exactly one scalar in ``__init__``:

    beta = float(self.config.get("beta", <sensible_default>))

Beta has three distinct roles. Do not conflate them:

1. **Within one replay or live episode:** beta is fixed. Route every behavioral
   threshold through one ``_schedule(beta) -> dict``. High beta means more width,
   deeper patience, and weaker pruning. Low beta means fewer probes, earlier
   stagnation stops, and stronger pruning. Never change beta from observations inside
   ``solve()``. Route recovery eligibility, reserve threshold, and waiting through
   the same schedule: high beta is more patient; low beta remains selective without
   treating one repairable failure as automatic closure.
2. **During offline evaluation:** eval sweeps a fixed beta grid. This measures whether
   the policy exposes a real attainment/work/parallelism trade-off; it is not online
   beta adaptation.
3. **When proposing the next policy version:** choose the baked-in default beta once,
   using evidence from earlier *live* cycles and their beta sweeps. That default will
   remain fixed throughout the next live exploration episode.

Keep all thresholds relative to the prefix; never use absolute score cutoffs.

Use the following cross-cycle default-beta rule. Read the most recent 2–3
**live** ``trace_pool/iter*/live_cycle_manifest.json`` sidecars (and ``_current``
when present) for each iteration's final best score and actual baked-in beta. Read
the matching archived ``beta_sweep.json`` values (``pareto.reward``, AUC, parallel
penalty, and the per-beta frontier). Scores alone do not establish that beta caused a
change, so always use both sources:

- live best is still improving: keep the prior default beta unless its sweep clearly
  shows a better nearby beta;
- live best has plateaued, and higher beta reaches higher attainment for a reasonable
  work/parallelism cost in the sweep: raise the default by a small step (about
  0.1–0.2, clamped to [0, 1]);
- a high default beta has already been tried through a plateau, and high-beta sweep
  points add work without higher attainment: lower it by a small step;
- history is insufficient or evidence conflicts: use a moderately exploratory default
  (about 0.6), rather than pretending the replay ceiling is a live stopping signal.

The beta sweep is non-degenerate only if beta changes the attainment/work trade-off.
It also reveals whether the policy batches. Do not select the default simply as the
smallest beta that reaches a frozen trace's known ceiling.

## Required next-cycle grid planning

Every proposed policy **must** implement this deterministic method:

    from see.policy.api import GridPlan, GridPlanningContext

    def plan_grid(self, context: GridPlanningContext) -> GridPlan:
        ...

This method runs **before** a new live grid is created. It does not make a
within-episode decision and must never inspect a current episode's outcomes.
It must always return a non-``None`` ``GridPlan``: do not inherit the template
stub and do not delegate grid choice to the runner's fallback. When history is
empty or insufficient, still return an explicit conservative bootstrap plan
derived from the context's fallback/hard-cap fields, with a factual reason.

``GridPlan(branch_count=W, refine_count=R)`` accepts arbitrary integers, not a
fixed set of presets. It creates branches ``0..W-1`` and attempts ``0..R``; ``R`` is
the number of refinements allowed after each root. The runner validates
``1 <= W <= context.hard_max_branch_count`` and
``0 <= R <= context.hard_max_refine_count``. In replay, a requested plan beyond the
frozen trace's ``context.trace_branch_count`` or ``context.trace_refine_count`` is
out of support and cannot earn replay reward.

Use only the prefix-safe facts in ``context``:

- ``history``: completed earlier live manifests, including prior planned/effective
  grids, actual opened width/depth, probe work, decision rounds, scores, and beta;
- fallback/hard caps and worker cap;
- replay structural support fields. Do not read raw trace outcomes or a current
  cycle result inside ``plan_grid``.

Choose width versus depth from evidence, not a default preference:

- many semantically distinct roots improve early while deeper refinements stall:
  increase width and reduce/hold depth;
- high gains arrive late on a small, repeatable set of directions: reduce/hold width
  and increase depth;
- all explored directions plateau after sufficient depth while meaningful direction
  classes remain uncovered: increase width;
- repeated hard, unrecoverable failures or strongly redundant directions: reduce
  width and depth conservatively;
- conflicting or insufficient history: return an explicit conservative bootstrap
  plan derived from the context, and state that evidence is insufficient.

Include a short, factual ``reason`` in every plan. ``plan_grid`` answers
how many directions to make available; the direction provider assigns those new
roots their directions, and ``solve`` still decides which legal roots/frontiers to
open, refine, prune, or stop. Do not choose roots merely because their branch id is
small. The runtime grid is the hard bound: controller thresholds may use less, but
can never create branches or attempts beyond the effective plan. Before finishing,
verify that the edited ``method.py`` contains an override of ``plan_grid`` that
returns ``GridPlan(branch_count=..., refine_count=..., reason=...)`` on every path.

## Learn from history without leaking outcomes

Earlier rounds are in ``{history_dir}/r####_*/``. Read their policy code and
``proposal_results/beta_sweep.json``. Start from a strong recent policy, retain
mechanisms that raised ``pareto.reward``, and make a concrete change when progress
stalls. A legacy AUC-only sweep is useful code history but is not numerically
comparable to the current reward. The baseline under ``{history_dir}/baseline/`` is
a parallel-refine floor to beat.

Each current-objective round also archives
``proposal_results/policy_execution_traces.jsonl``: one replay episode per
``(frozen trace, beta)``. Use it to diagnose general behavior — serial batches,
premature stops, over-pruning, or wasted probes — from the prefix state, selected
batch, and revealed outcomes at each decision round. It is **between-round feedback
only**: never read it inside ``solve()``, and never copy a trace-specific branch,
cell id, score, or target into policy logic.

``{trace_pool}``, if present, may be read only outside ``solve()``. Prefer the
``live_cycle_manifest.json`` sidecars over raw replay outcomes for the per-iteration
live trend. Never copy trace scores, targets, or cell ids into policy logic.

## Deliverable

Write a complete adaptive policy in ``{method_file}``. Include a short module
docstring describing its prefix signals, batch rule, beta schedule, default-beta
rationale, grid-planning rule (if implemented), and safeguards against
over-pruning, over-stopping, permanent starvation after repairable failures, and
serial probes. Before finishing, verify trajectory-based ranking, the stated
success semantics, non-automatic zero-valid closure, deterministic recovery
competition, and portfolio-level stop.
\end{lstlisting}

\section{Discovered Programs}
\subsection{Lasso Regularization Path}
\label{app:discovered-lasso}
We provide the complete implementation of the Lasso-path solver discovered by
\textsc{Dream-RSI}. As discussed in Section~\ref{subsec:lasso}, the solver combines
strong-rule screening with adaptive Cauchy--Schwarz KKT pruning, disjoint
active-set bookkeeping, lazy Gram-matrix construction, and hardware-aware
optimizations.

\lstinputlisting[
    style=appendixcode,
    language=Python,
    caption={Complete Lasso-path solver discovered by \textsc{Dream-RSI}.},
    label={lst:discovered-lasso}
]{appendix/discovered_programs/lasso/code.cpp}

\subsection{ConvDiv}
\label{app:discovered-convdiv}
\begin{lstlisting}[
    style=appendixcode,
    language=Python,
    caption={Complete ConvDiv Kernel discovered by \textsc{Dream-RSI}.},
    label={lst:discovered-convdiv}
]
# EVOLVE-BLOCK-START
import torch
import torch.nn as nn
import triton
import triton.language as tl
import importlib.util
import os
import sys

def generate_kernel_code():
    code = """
import triton
import triton.language as tl

@triton.jit
def fused_conv_pool_kernel_optimized(
    X_ptr, W_ptr, Out_ptr,
    B: tl.constexpr
):
    pid_h = tl.program_id(0)
    pid_w = tl.program_id(1)
    pid_b = tl.program_id(2)

    h_conv_0 = 2 * pid_h

    w_start = pid_w * 32
    w_out_idx = w_start + tl.arange(0, 32)
    # W is physically 64, but kernel is 3, so valid output is 62.
    mask_w = (w_out_idx < 62)[None, :]

    # 99.6% of the execution takes the safe path
    is_safe = (pid_w == 0) or (pid_b != B - 1)

    w_ptrs_base = W_ptr + tl.arange(0, 16)[:, None] * 80 + tl.arange(0, 16)[None, :]

"""
    for i in range(14):
        code += f"    acc_{i}_0 = tl.zeros((16, 32), dtype=tl.float32)\n"
        code += f"    acc_{i}_1 = tl.zeros((16, 32), dtype=tl.float32)\n"

    code += """
    X_spatial_base_0 = pid_b * 524288 + h_conv_0 * 64 + w_out_idx
    X_base_ptr_0 = X_ptr + X_spatial_base_0[None, :]

    for k_start in range(0, 80, 16):
        k_idx = k_start + tl.arange(0, 16)
        k_idx_clamped = tl.minimum(k_idx, 71)
        ic = k_idx_clamped // 9
        kh = (k_idx_clamped % 9) // 3
        kw = k_idx_clamped % 3
        k_off_vec = ic * 65536 + kh * 64 + kw

        x_ptrs_0 = X_base_ptr_0 + k_off_vec[:, None]

        w_ptrs = w_ptrs_base + k_start
        w0 = tl.load(w_ptrs)
        w1 = tl.load(w_ptrs + 1280)
        w2 = tl.load(w_ptrs + 2560)

        # --- H = 0 Pass ---
        # Branch intentionally kept inside loop to prevent cp.async and preserve SM occupancy
        if is_safe:
"""
    for d in range(16):
        code += f"            x{d} = tl.load(x_ptrs_0 + {d * 4096}).to(tl.float16)\n"
    code += "        else:\n"
    for d in range(16):
        code += f"            x{d} = tl.load(x_ptrs_0 + {d * 4096}, mask=mask_w, other=0.0).to(tl.float16)\n"

    code += "\n        # Perfect Wide-ILP sequences (14-instruction dependency gap)\n"
    for w_idx in range(3):
        for d in range(14):
            code += f"        acc_{d}_0 = tl.dot(w{w_idx}, x{d + w_idx}, acc_{d}_0)\n"

    code += """
        # --- H = 1 Pass (Registers completely reused) ---
        if is_safe:
"""
    for d in range(16):
        code += f"            x{d} = tl.load(x_ptrs_0 + {64 + d * 4096}).to(tl.float16)\n"
    code += "        else:\n"
    for d in range(16):
        code += f"            x{d} = tl.load(x_ptrs_0 + {64 + d * 4096}, mask=mask_w, other=0.0).to(tl.float16)\n"

    code += "\n"
    for w_idx in range(3):
        for d in range(14):
            code += f"        acc_{d}_1 = tl.dot(w{w_idx}, x{d + w_idx}, acc_{d}_1)\n"

    code += "\n    # Pool D and H passes simultaneously\n"
    for d in range(7):
        code += f"    max_h_d{d} = tl.maximum(tl.maximum(acc_{d*2}_0, acc_{d*2+1}_0), tl.maximum(acc_{d*2}_1, acc_{d*2+1}_1))\n"

    code += "\n    # Deferred W-pooling outside the loop on 16x32 registers\n"
    for d in range(7):
        code += f"    max_pool_d{d} = tl.max(tl.reshape(max_h_d{d}, (16, 16, 2)), axis=2)\n"

    code += """
    # Balanced Tree Reduction
    s0 = max_pool_d0 + max_pool_d1
    s1 = max_pool_d2 + max_pool_d3
    s2 = max_pool_d4 + max_pool_d5
    s01 = s0 + s1
    s26 = s2 + max_pool_d6
    sum_all = s01 + s26

    # Deferred W-Masking: Zero out garbage from unpredicated reads on the right boundary.
    if pid_w == 1:
        w_pool_idx = tl.arange(0, 16)[None, :]
        sum_all = tl.where(w_pool_idx < 15, sum_all, 0.0)

    total_sum = tl.sum(sum_all, axis=1)

    # Unscaled block sum
    out_val = tl.sum(total_sum, axis=0)

    out_idx = pid_b * 62 + pid_w * 31 + pid_h
    tl.store(Out_ptr + out_idx, out_val)

@triton.jit
def reduce_scale_bias_kernel(
    Out_buf_ptr,
    Out_ptr,
    scale,
    bias,
    B: tl.constexpr
):
    pid_b = tl.program_id(0)
    if pid_b >= B:
        return

    offsets = tl.arange(0, 64)
    mask = offsets < 62
    vals = tl.load(Out_buf_ptr + pid_b * 62 + offsets, mask=mask, other=0.0)

    sum_val = tl.sum(vals)
    final_val = sum_val * scale + bias

    tl.store(Out_ptr + pid_b, final_val)
"""
    return code

_KERNEL_CODE = generate_kernel_code()

_kernel_file = "/tmp/fused_kernel_kb8_fp16_optimized_v4.py"
with open(_kernel_file, "w") as f:
    f.write(_KERNEL_CODE)

spec = importlib.util.spec_from_file_location("fused_kernel_kb8_fp16_optimized_v4", _kernel_file)
mod = importlib.util.module_from_spec(spec)
sys.modules["fused_kernel_kb8_fp16_optimized_v4"] = mod
spec.loader.exec_module(mod)
fused_conv_pool_kernel_optimized = mod.fused_conv_pool_kernel_optimized
reduce_scale_bias_kernel = mod.reduce_scale_bias_kernel


class ModelNew(nn.Module):
    def __init__(self, in_channels, out_channels, kernel_size, divisor, pool_size, bias_shape, sum_dim):
        super(ModelNew, self).__init__()
        self.conv = nn.Conv3d(in_channels, out_channels, kernel_size)
        self.divisor = divisor
        self.max_pool = nn.MaxPool3d(pool_size)
        self.global_avg_pool = nn.AdaptiveAvgPool3d((1, 1, 1))
        self.bias = nn.Parameter(torch.randn(bias_shape))
        self.sum_dim = sum_dim

        self._cached_W = None
        self._cached_scale = None
        self._total_bias = None
        self._cached_Out_buf = None

    def forward(self, x):
        B = x.shape[0]
        # Guard: Ensure input is exactly the expected shape
        if x.shape[1:] != (8, 16, 64, 64) or not x.is_cuda or not x.is_contiguous() or self.sum_dim not in (1, -4):
            x = self.conv(x)
            x = x / self.divisor
            x = self.max_pool(x)
            x = self.global_avg_pool(x)
            x = x + self.bias
            x = torch.sum(x, dim=self.sum_dim)
            return x

        device = x.device

        if self._cached_W is None:
            W = self.conv.weight
            W_permuted = W.permute(2, 0, 1, 3, 4).contiguous()
            W_flat = W_permuted.view(3, 16, 72)

            W_padded = torch.zeros((3, 16, 80), dtype=torch.float16, device=device)
            W_padded[:, :, :72] = W_flat.to(torch.float16)
            self._cached_W = W_padded

            # The spatial dimensions before global average pooling are:
            # Depth = 7, Height = 31, Width = 31.
            # 7 * 31 * 31 = 6727.
            self._cached_scale = float(1.0 / (6727.0 * self.divisor))
            bias_per_channel = (self.conv.bias / self.divisor + self.bias.view(-1)).to(torch.float32)
            self._total_bias = float(bias_per_channel.sum().item())

        if self._cached_Out_buf is None or self._cached_Out_buf.shape[0] != B:
            self._cached_Out_buf = torch.empty((B, 62), dtype=torch.float32, device=device)

        Out_buf = self._cached_Out_buf
        Out = torch.empty((B, 1, 1, 1), dtype=torch.float32, device=device)

        # Launching on 2D grid packed perfectly to valid H values: (31, 2, B)
        grid = (31, 2, B)
        # Using num_stages=4 to preserve original execution profile
        fused_conv_pool_kernel_optimized[grid](
            x, self._cached_W, Out_buf,
            B=B,
            num_warps=4,
            num_stages=4
        )

        reduce_grid = (B,)
        reduce_scale_bias_kernel[reduce_grid](
            Out_buf, Out,
            self._cached_scale, self._total_bias,
            B=B,
            num_warps=2
        )

        return Out
# EVOLVE-BLOCK-END
\end{lstlisting}

\end{document}